\documentclass{article}
\usepackage{microtype}
\usepackage{graphicx}
\usepackage{svg}
\usepackage{subfigure}
\usepackage{booktabs} % for professional tables
\usepackage{hyperref}
\usepackage{color}
\usepackage[normalem]{ulem}
\usepackage[accepted]{icml2024}
\usepackage{amssymb}
\usepackage{mathtools}
\usepackage{amsthm}
\usepackage{enumitem}
\usepackage{dsfont}
\usepackage{enumitem}
\usepackage{multirow}

\newcommand{\bamo}{BA-2motifs}

\newcommand{\benz}{Benzene}
\newcommand{\mutag}{MUTAG}

\newcommand{\fluo}{Fluoride} %-Carbonyl
\newcommand{\alk}{Alkane} %-Carbonyl
\newcommand{\dd}{D\&D}
\newcommand{\prot}{PROTEINS}

\theoremstyle{plain}
\newtheorem{theorem}{Theorem}[section]
\newtheorem{proposition}[theorem]{Proposition}

\newtheorem{definition}[theorem]{Definition}

\newtheorem{Example}[theorem]{Example}
\theoremstyle{remark}

\usepackage[textsize=tiny]{todonotes}
\usepackage{amsmath,amsfonts,bm}

\def\eqref#1{equation~\ref{#1}}
\def\1{\bm{1}}

\def\mA{{\bm{A}}}

\def\mX{{\bm{X}}}

\DeclareMathAlphabet{\mathsfit}{\encodingdefault}{\sfdefault}{m}{sl}
\SetMathAlphabet{\mathsfit}{bold}{\encodingdefault}{\sfdefault}{bx}{n}

\def\sR{{\mathbb{R}}}

\DeclareMathOperator*{\argmin}{arg\,min}

\newcommand{\stitle}[1]{\vspace{1ex}\noindent{\bf #1}}

\newcommand{\ms}[2]{{#1\tiny{$\pm$#2}}}

\icmltitlerunning{PAC Learnability under Explanation-Preserving Graph Perturbations}

\begin{document}

\twocolumn[
\icmltitle{PAC Learnability under Explanation-Preserving Graph Perturbations}

\icmlsetsymbol{equal}{*}

\begin{icmlauthorlist}
\icmlauthor{Xu Zheng}{equal,fiu}
\icmlauthor{Farhad Shirani}{equal,fiu}
\icmlauthor{Tianchun Wang}{psu}
\icmlauthor{Shouwei Gao}{fiu} 
\icmlauthor{Wenqian Dong}{fiu}
\icmlauthor{Wei Cheng}{nec}
\icmlauthor{Dongsheng Luo}{fiu}

\end{icmlauthorlist}

\icmlaffiliation{fiu}{Florida International University}
\icmlaffiliation{psu}{Pennsylvania State University}
\icmlaffiliation{nec}{NEC Lab America}

\icmlkeywords{Graph Learning, PAC Learnability}

\vskip 0.3in
]

\begin{abstract}
Graphical models capture relations between entities in a wide range of  applications including social  networks, biology, and natural language processing, among others. Graph neural networks (GNN) are neural models that operate over graphs, enabling the model to leverage the complex relationships
and dependencies in graph-structured data. A graph explanation is a subgraph which is an `almost sufficient' statistic of the input graph with respect to its classification label. Consequently, the classification label is invariant, with high probability, to perturbations of graph edges not belonging to its explanation subgraph. This work considers two methods for leveraging such perturbation invariances in the design and training of GNNs. First, explanation-assisted learning rules are considered. It is shown that the sample complexity of explanation-assisted learning can be arbitrarily smaller than explanation-agnostic learning. Next, explanation-assisted data augmentation is considered, where the training set is enlarged by artificially producing new training samples via perturbation of the non-explanation edges in the original training set. It is shown that such data augmentation methods may improve performance if the augmented data is in-distribution, however, it may also
lead to worse sample complexity compared to explanation-agnostic learning rules if the augmented data is out-of-distribution. Extensive empirical evaluations are provided to verify the theoretical analysis. 
\end{abstract}

\printAffiliationsAndNotice{\icmlEqualContribution} % otherwise use the standard text.

\section{Introduction}
\label{sec:intro}
Graphical models represent relations between entities in a wide range of  applications such as social  networks, biology, transportation systems, natural language processing, and computer vision \cite{newman2018networks}. 
Traditional machine learning approaches
often struggle to effectively leverage the rich relational information encoded in graphs. To address this, and inspired by conventional deep learning methods, various graph neural network (GNN) architectures have been developed, such as methods based on convolutional neural networks~\cite{defferrard2016convolutional,kipf2017semisupervised}, recurrent neural networks~\cite{li2016gated,ruiz2020gated},  and transformers~\cite{yun2019graph, rong2020self}. 
The application of machine learning solutions in critical domains such as autonomous vehicles, medical diagnostics, and financial systems has created an urgent need for explainable learning methods \cite{ying2019gnnexplainer,luo2020parameterized,yuan2022explainability,yuan2021explainability}. At a high level, explainability can be interpreted as identifying a subset of the input that significantly \textit{influences} the model's output. In the context of GNNs, this often translates to identifying an influential subgraph of the input graph. Recent works have proposed gradients-based~\cite{pope2019explainability,baldassarre2019explainability}, perturbed-predictions-based~\cite{ying2019gnnexplainer,luo2020parameterized,yuan2021explainability,shan2021reinforcement}, and decomposition-based~ \cite{baldassarre2019explainability} methods for GNN explainability.

The notion of subgraph explainability is based on the intuitive assumption that the presence of certain structural patterns or motifs within the input graph plays a critical role in the model's decision-making process \cite{ying2019gnnexplainer,luo2020parameterized,yuan2021explainability,shan2021reinforcement}.
  Consequently, prior works have defined a graph explanation as a subgraph which is an \textit{almost sufficient} statistic of the input graph with respect to the classification label. The design of many of the widely recognized GNN explanation algorithms such as  GNNExplainer \cite{ying2019gnnexplainer}, PGExaplainer \cite{luo2020parameterized}, and GSAT \cite{miao2023interpretable}, as well as fidelity measures such as $Fid_+$, $Fid_-$, and $Fid_{\Delta}$ \cite{pope2019explainability,yuan2022explainability} are based on the assumption that the presence or absence of the explanation subgraph in the input determines the output label with high accuracy, and hence perturbations of the non-explanation edges --- graph edges which do not belong to the explanation subgraph --- does not change the output label with high probability. 
  
  The invariance of the GNN output to perturbations of non-explanation edges resembles the transformation invariances observed in various learning tasks on non-graphical data such as invariance to scaling and rotation in image classification tasks \cite{cohen2016group,bloem2020probabilistic,chen2020group,shao2022theory}. Such transformation invariances have been leveraged in both network architectural design and data augmentation in the context of image classification \cite{krizhevsky2012imagenet,he2016deep}, electroencephalographic (EEG) classification \cite{krell2017rotational}, and radio modulation classification \cite{huangdata}, among other applications.

  Analogous to the aforementioned prior works, which leverage various types of transformation invariances in non-graph learning applications to design learning mechanisms and data augmentation methods, in this work, we wish to study how one can leverage the invariance to perturbations of non-explanation edges for GNN architecture design and data augmentation.  Particularly, we consider learning scenarios, where, in addition to labeled graph samples, each training sample is accompanied by its ground-truth explanation subgraph. Such ground-truth explanations may be produced at the time that the training data is compiled. For example, in a dataset of labeled radiology scans, the most informative sections of each scan could be identified by the contributing physicians during the compilation phase of the training dataset. Alternatively, an estimate of the explanation can be produced by joint training of a GNN explainer and classifier using the original (unexplained) training data. Considering such scenarios, we introduce \textit{explanation-assisted} learning rules and data augmentation methods. 

  Our main contributions are summarized in the following:
  \begin{itemize}[itemsep=1.5pt,topsep=0pt,parsep=0pt,leftmargin=*]
      \item To provide a rigorous formulation of the explanation-assisted graph learning problem and the associated sample complexity. 
      \item To introduce the explanation-assisted empirical risk minimization (EA-ERM) learning rule, and derive its sample complexity, and to prove the optimality of the EA-ERM in terms of sample complexity. (Theorem \ref{th:1})
      \item To show that the EA-ERM sample complexity can be arbitrarily smaller than the (explanation-agnostic) ERM sample complexity. (Example \ref{ex:arb})
      \item To introduce explanation-assisted data augmentation mechanisms, and to provide a theoretical justification, along with an example, showing that while in some scenario such data augmentation mechanisms may improve performance, there also exist scenarios where they lead to worse sample complexity even compared to that of the explanation-agnostic learners. (Example \ref{ex:DAERM})
      \item To provide an implementable class of explanation-assisted GNN mechanisms by building on the intuition from our theoretical analaysis of EA-ERM. (Section \ref{sec:learn})
      \item To provide empirical simulations verifying the improved performance of our explanation-assisted GNN architectures when the necessary conditions in our theoretical derivations are satisfied; and to provide empirical simulations illustrating the diminished performance in scenarios not satisfying the necessary conditions. (Section \ref{sec:sim})
  \end{itemize}

\section{Preliminaries}
\label{sec:pre}
This section introduces the notation and some of the necessary background concepts used in the rest of the paper.
\subsection{The Classification Problem}
A random labeled graph $G$ is parametrized by i) a vertex\footnote{We use node and vertex interchangeably.} set $\mathcal{V}=\{v_1,v_2,\cdots,v_n\}$, where $n\in \mathbb{N}$, ii) an edge set $\mathcal{E}\subseteq \mathcal{V}\times \mathcal{V}$, iii)
a feature matrix $\mX \in \sR^{n\times d}$, where  the $i$th row $\mX_i$ is the feature vector associated with $v_i$ and $d$ is the dimension of the feature vectors, iv) an adjacency matrix $\mA \in \{0,1\}^{n\times n}$, where $A_{i,j}=\mathds{1}((v_i,v_j)\in \mathcal{E})$,  and v) a 
label $Y\in \mathcal{Y}$, where $\mathcal{Y}$ is a finite set. The graph parameters $(Y,\mA,\mX)$
are generated based on the joint distribution $P_{Y,\mA,\mX}$. The notation $P_G$ and $P_{Y,\mA,\mX}$ are used interchangeably. 
For a labeled graph $G=(\mathcal{V},\mathcal{E};Y,\mX,\mA)$, the corresponding graph without label is denoted as $\overline{G}=(\mathcal{V},\mathcal{E};\mX,\mA)$. The induced marginal distribution of $\overline{G}$ is $P_{\overline{G}}$ and its support is $\overline{\mathcal{G}}$. 
A classification scenario is completely characterized by $P_G$, consequently, in the rest of the paper, we refer to $P_G$ as  \textit{the classification problem}.

\begin{definition}[\textbf{Graph Classification}]
A graph classifier for a classification problem $P_G$ is a function $f  :\overline{\mathcal{G}} \to \mathcal{Y}$. 
Given $\epsilon\in [0,1]$, the classifier is called $\epsilon$-accurate if  $P_{G}(f(\overline{G}) \neq  Y)\leq \epsilon$.
\end{definition}

\subsection{Generic Learning Rules and ERM}
A training set $\mathcal{T}$ is a collection of labeled  graphs. The elements of the training set are generated independently and according to $P_G$.
A learning rule is a procedure which takes the training set $\mathcal{T}$ as input, and outputs a graph classifier belonging to an underlying hypothesis class $\mathcal{H}$. 
\begin{definition}[\textbf{Generic Learning Rule}]
\label{def:learn}
Let the hypothesis class $\mathcal{H}$ be a collection of graph classifiers. 
A generic learning rule $\mathsf{L}=(L_t)_{t\in \mathbb{N}}$ consists of a family of mappings $L_t:\mathcal{T}_t\mapsto f(\cdot)$,
where the input $\mathcal{T}_t=\{(\overline{G}_i,Y_i),i\in [t]\}$ is called the \textit{training set}, 
and the output $f:\overline{\mathcal{G}} \to \mathcal{Y}$ is a graph classifier belonging to the hypothesis class $\mathcal{H}$. 
\end{definition}

The empirical risk minimization (ERM) learning rule is a subclass of generic learning rules defined below.

\begin{definition}[\textbf{Empirical Risk Minimization}]
 Given a hypothesis class $\mathcal{H}$ and family of training sets $\mathcal{T}_t=\{(\overline{G}_i,Y_i),i\in [t]\}, t\in \mathbb{N}$, the learning rule $\mathsf{L}_{{ERM}}=(L_{ERM,t})_{t\in \mathbb{N}}$ is defined as:
\begin{align*}
    L_{ERM,t}(\mathcal{T}_t)\triangleq \argmin_{f(\cdot)\in \mathcal{H}} \frac{1}{t}\sum_{i=1}^t \mathds{1}(f(\overline{G}_i)
\neq Y_i),\quad t\in \mathbb{N}.
\end{align*}
\end{definition}

\subsection{GNN Explainability}

The objective in instance-level GNN explanability is to take a graph $\overline{G}$ as input, and construct a subgraph $\overline{G}_{exp}$ which is an (almost) sufficient statistic of the input graph with respect to its label. We adapt the formalization introduced in \cite{zheng2023robust} which is summarized in the following.
%This is formalized in the following.

\begin{definition}[\textbf{Explanation Function}]
\label{def:exp}
Given a classification problem $P_G$, an explanation function (explainer) is a mapping
$\Psi: \overline{\mathcal{G}}\to 2^{\mathcal{V}}\times 2^{\mathcal{E}}$, such that
\begin{align*}
    &\Psi(\overline{G})= (\mathcal{V}_{exp},\mathcal{E}_{exp}), \overline{G}\in \overline{\mathcal{G}}, 
\end{align*}
where $\mathcal{V}_{exp}\subseteq \mathcal{V}$, $\mathcal{E}_{exp}=(\mathcal{V}_{exp}\times \mathcal{V}_{exp})\cap \mathcal{E}$, and $\mathcal{V}$ and $\mathcal{E}$ are the vertex set and edge set of $\overline{G}$, respectively. 
For a given pair of parameters $\kappa\in [0,1]$ and $s\in \mathbb{N}$, 
the explainer $\Psi(\cdot)$ is called an $(s,\kappa)$-explainer if:
\begin{align*}
&  i) \quad  I(Y;\overline{G}| \mathds{1}_{\Psi(\overline{G})}) \leq \kappa,
 \qquad \qquad ii) \quad \mathbb{E}_{G}(|\mathcal{E}_{exp}|)\leq s,
\end{align*}
where we have defined 
\begin{align*}
&I(Y\!;\! \overline{G}|\mathds{1}_{\Psi(\overline{G})})\triangleq\!\!
\sum_{{g}_{exp}}\!\!P_{\Psi(\overline{G})}(g_{exp})\sum_{y,\overline{g}} P_{Y,\overline{G}}(y, \overline{g}| g_{exp}\!\subseteq\! \overline{G})
\\& \qquad \qquad \qquad \times \log\frac{P_{Y,\overline{G}}(y, \overline{g}| g_{exp}\subseteq \overline{G})}{P_{Y}(y|  g_{exp}\subseteq \overline{G})P_{\overline{G}}(\overline{g}|  g_{exp}\subseteq \overline{G})}.
\end{align*}
If such an explainer exists, the classification problem is said to be $(s,\kappa)$-explainable.
\end{definition}

To keep the analysis tractable, for all explanation functions considered in our theoretical analysis, we assume that:
    \begin{align}
    \label{eq:cond}
       \text{Condition 1: }  \forall \overline{g},\overline{g}'\in \mathcal{\overline{G}}\!:\! \Psi(\overline{g})\!\subseteq \overline{g}'\! \Rightarrow \!\Psi(\overline{g}')=\Psi(\overline{g}).
         %\quad \forall \overline{g}_{exp}\neq \phi: \exists \overline{g}: \Psi(\overline{g})=g_{exp} \iff \forall \overline{g}: \overline{g}_{exp}\subseteq \overline{g} \to \Psi(\overline{g})=\overline{g}_{exp}.
    \end{align}
This implies that $I(Y;\overline{G}|\mathds{1}_{\Psi(\overline{G})})= I(Y;\overline{G}|\Psi(\overline{G}))$. 
 The condition holds for the ground-truth explanation in various datasets studied in the explainability literature such as BA-2motifs, Tree-Cycles, Tree-Grid, and MUTAG datasets. For a more complete background discussion of Definition \ref{def:exp} and Condition 1, and its connections with other widely used explainability measures such as graph information bottleneck (GIB), please refer to \cite{zheng2023robust}. 

 \subsection{Explanation-Assisted Learning Rules}
 We define explanation-assisted learning rules and their associated sample complexity as follows.

\begin{definition}[\textbf{Explanation-Assisted Learning Rule}]
\label{def:EA-learn}
Given a hypothesis class $\mathcal{H}$, an explanation-assisted learning rule $\mathsf{L}_{EA}=(L_{EA,t})_{t\in \mathbb{N}}$ consists of a family of mappings $L_{EA,t}:(\mathcal{T}_t,\Psi_{|\mathcal{T}_t}(\cdot))\mapsto f(\cdot)$,
where $\mathcal{T}_t, t\in \mathbb{N}$ is the training set, $\Psi(\cdot)$ is an explanation function, and $\Psi_{|\mathcal{T}_t}(\cdot)$ is its restriction to the training set. 
\end{definition}

To define the sample complexity of explanation-assisted learning rules over \textit{explainable} classification problems, we first introduce the class of perturbation-invariant classification problems. In the subsequent sections, we quantify the relation between perturbation-invariance and explainability (see Proposition \ref{Prop:1}).
\begin{definition}[\textbf{Perturbation-Invariant Classification Problem}]
\label{def:pert_aware}
Given an explainer $\Psi(\cdot)$ and parameters $\zeta\in [0,1]$, a classification problem $P_G$ is $(\Psi,\zeta)$-invariant if 
% \begin{align*}
%     P(Y_G\neq Y_{\overline{G}'}| \overline{G}'\in \mathcal{S}^{\gamma}(\overline{G}))\leq \zeta,
% \end{align*}
\begin{align*}
&\sum_{g_{exp}}P(\Psi(\overline{G})=g_{exp})\times \\&\qquad P(Y_{\overline{G}'}\neq Y_{\overline{G}''}|\Psi(\overline{G}')=\Psi(\overline{G}'') =g_{exp})\leq \zeta,
\end{align*}
where $Y_{\overline{G}'}$ and $Y_{\overline{G}''}$ are the labels associated with $\overline{G}'$ and $\overline{G}''$, respectively, and  the labeled graphs $\overline{G},\overline{G}',\overline{G}''$ are generated independently and  according to $P_{\overline{G}}$

\end{definition}

As shown in our empirical analysis, the perturbation-invariant property can be empirically observed in many of the graph classification datasets considered in the literature, such as the BA2-Motifs dataset \cite{luo2020parameterized}, where the graph label is completely determined by the presence of specific motifs, and does not change with perturbations of edges in the graph that are not part of the motifs. Similarly, in the Mutag dataset, labels indicating mutagenicity are associated with various motifs such as aromatic nitro groups, polycyclic aromatic hydrocarbons, and alkylating agents \cite{debnath1991structure}. We define sample complexity of explanation-assisted learning rules over perturbation invariant classification problems in the following.

 \begin{definition}[\textbf{Explanation-Assisted Sample Complexity}]
\label{def:EASC}
For any $\epsilon,\delta,\zeta\in (0,1)$, the sample complexity of $(\epsilon,\delta,\zeta)$-PAC learning of $\mathcal{H}$ with respect to explanation function $\Psi(\cdot)$, denoted by $m_{EA}(\epsilon,\delta,\zeta;\mathcal{H},\Psi)$, is defined as the smallest $m \in \mathbb{N}$ for which there exists an explanation-assisted learning rule $\mathsf{L}$ such that, for every $(\Psi,\zeta)$-invariant classification problem $P_G$, we have:
\begin{align*}
    P\left(err_{P_G}\left(\mathsf{L}(\mathcal{T})\right)\leq \inf_{f\in \mathcal{H}} err_{P_G}(f) +\epsilon\right)\geq 1-\delta,
\end{align*}
where we have defined $err_{P_G}(f)$ as the statistical error of $f(\cdot)$ on $P_G$, and probability is evaluated with respect to the training set $\mathcal{T}$. If no such $m$ exists, the we say the sample complexity is infinite.
\end{definition}

 \section{PAC Learnability of Explanation-Assisted Learners}
\label{sec:PAC}
In this section, we introduce the explanation-assisted ERM (EA-ERM),
characterize its sample complexity, and prove its optimality.  We provide an example, where this sample complexity can be arbitrarily smaller than the standard explanation-agnostic sample complexity achieved by ERM. We conclude that for  explainable classification problems there may be significant benefits in using explanation-assisted learning rules, in terms of size of the training set required to achieve a specific error probability. This is further verified via empirical analysis in the subsequent sections.  

As a first step, the following proposition provides sufficient conditions for perturbation-invariance of a classification problem in terms of its explainability parameters and its Bayes error probability.

\begin{proposition}[\textbf{Perturbation Invariance and Explainability}]
\label{Prop:1}
Let $\zeta\in [0,1]$. There exist $ \overline{\kappa},\overline{\epsilon}>0, s\in \mathbb{N}$, such that for all $ \kappa\leq \overline{\kappa}$ and $\epsilon\leq \overline{\epsilon}$, any $(\kappa,s)$-explainable classification problem $P_G$ with Bayes error $\epsilon$ is $(\Psi,\zeta)$-invariant, where $\Psi(
\cdot)$ is a $(\kappa,s)$-explanation function for $P_G$.
\end{proposition}
The proof is provided in Appendix \ref{App:Prop:1}.

 The notion of perturbation invariance is analogous to transformation invariances, such as rotation and scaling invariances, observed in image classification. Prior works on sequential data have shown that in those contexts invariance-aware learning rules can achieve improved sample complexity, e.g.,  \cite{shao2022theory}. 
Building on this, we define EA-ERM as follows.

\begin{definition}[\textbf{Explanation-Assisted ERM (EA-ERM)}]
\label{def:EAERM}
Given a hypothesis class $\mathcal{H}$, family of training sets $\mathcal{T}_t, t\in \mathbb{N}$, and explanation function $\Psi(\cdot)$, the learning rule $\mathsf{L}_{\text{EA-ERM}}= (L_{\text{EA-ERM},t})_{t\in\mathbb{N}}$ is defined as:
\begin{align}
 &\nonumber L_{\text{EA-ERM},t}\triangleq \widetilde{f}_{t}(\cdot), \quad t\in \mathbb{N}
 \\&    \widetilde{f}_t(\overline{G})\!\triangleq \!
    \begin{cases}
        Y_{exp}\quad & \exists i\in [t]\!:\! \Psi(\overline{G}_i)\subseteq \overline{G},\\
        f_t(\overline{G}) & \text{Otherwise}
    \end{cases},\label{eq:second}
 \\&\nonumber f_{t}(\cdot)\triangleq  \mathsf{L}_{ERM,t}(\mathcal{T}_t),
\end{align}
where $Y_{exp}$ is chosen randomly and uniformly from the set $\{Y_i|  \Psi(\overline{G}_i)\subseteq \overline{G}, i\in [t]\}$.
\end{definition}

Note that Definition \ref{def:EAERM} implies a two-step learning procedure. First, 
given a training set $\mathcal{T}_t$, a classifier $f_t(\cdot)$ is trained by applying  the ERM learning rule $\mathsf{L}_{ERM,t}$. Then, $\widetilde{f}_t(\cdot)$ is constructed from $f_t(\cdot)$ using \eqref{eq:second}. This step ensures that the output of the classifier is the same for all explanation-preserving perturbations of the training samples. The second step was shown to be necessary to achieve improved sample complexity in \cite{shao2022theory} in the context of transformation invariances.

\begin{definition}[\textbf{Explanation-Assisted VC Dimension}]
   Given an explanation function $\Psi(\cdot)$ and hypothesis class $\mathcal{H}$, the explanation-assisted VC dimension  $VC_{EA}(\mathcal{H},\Psi)$ is defined as the largest integer $k$ for which there exists a collections of labeled graphs $\mathcal{G}=\{\overline{g}_1,\overline{g}_2,\cdots,\overline{g}_k\}$ such that $\Psi(\overline{g}_i)\neq \Psi(\overline{g}_j)$ for all $i\neq j$, and every labeling of $\mathcal{G}$ is realized by the hypothesis class $\mathcal{H}$. 
\end{definition}
Let us define $\mathbf{I}(\overline{G})=\overline{G}$ as the identity function. We call $VC(\mathcal{H})\triangleq VC_{EA}(\mathcal{H},\mathbf{I})$ the \textit{standard} VC dimension as it aligns with the notion of VC dimension considered in traditional PAC learnability analysis. The following provides a simple example, in which the standard VC dimension, $VC(\mathcal{H})$, can be arbitrarily larger than the explanation-assisted VC dimension $VC_{EA}(\mathcal{H},\Psi(\cdot))$.

\begin{Example}
\label{ex:arb}
Let $C_i, i\in \mathbb{N}$ denote the single-cycle graph with $i$ vertices, where the vertex set is $\mathcal{V}_i=[i]$ and the edge set is $\mathcal{E}_i= \{(j,j+1), j\in [i-1]\}\cup \{(1,i)\}$. We construct a binary classification problem as follows. Let the graphs associated with label zero belong to the collection $\mathcal{B}_0=\{C_i\cup C_3, i>5\}$ and those associated with label one belong to $\mathcal{B}_2=\{C_i\cup C_4, i>5\}$. Let 
\[\Psi(\overline{G})=
\begin{cases}
 C_3\qquad &\text{ if } \overline{G}\in \mathcal{B}_0\\
 C_4&\text{ otherwise}
\end{cases}.
\]
Clearly, $\zeta=0$ in this case. 
Let $P_Y(\cdot)$ be Bernoulli with parameter $\frac{1}{2}$, so that the two labels are equally likely, and assume that for a given label $Y=y$, the graphs belonging to $\mathcal{B}_y$ are equally likely, i.e., $P_{\overline{G}|Y}(\cdot|y)$ is uniform. Let $\mathcal{H}$ consist of all possible classifiers on the set $\mathcal{B}_0\cup \mathcal{B}_1$. So that $VC(\mathcal{H})=\infty$. It is straightforward to see that $VC(\mathcal{H},\Psi(\cdot))=2$ since there are only two explanation graphs, namely $C_3$ and $C_4$.
\end{Example}

Next, we show that explanation-assisted learning rules achieve sample complexity equal to $VC_{EA}(\mathcal{H},\Psi(\cdot))$  as opposed to $VC(\mathcal{H})$ achieved by generic learning rules. 

\begin{theorem}[\textbf{Sample Complexity of Explainable Tasks}]
\label{th:1}
Let $\epsilon,\delta\in (0,1)$ and $\zeta\leq \frac{\epsilon}{32}$. For any hypothesis class $\mathcal{H}$ and explanation function $\Psi(\cdot)$, the following holds:
\begin{align*}
m_{EA}(\epsilon,\delta,\zeta;\mathcal{H},\Psi) = {O}\left(\frac{d}{\epsilon^2}\log^2{d}+\frac{1}{\epsilon^2}ln(\frac{1}{\delta})\right),   
\end{align*}
where we have defined $d\triangleq VC_{EA}(\mathcal{H},\Psi(\cdot))$.
\end{theorem}
The proof is provided in Appendix \ref{App:th:1}.

\section{PAC Learnability of Explanation-Assisted Dataset Augmentation}
\label{sec:aug}
In the previous section, we showed that explanation-assisted learning rules outperform generic explanation-agnostic learning rules in terms of sample complexity. This suggests that if explanation subgraphs associated with each of the training samples are accessible, then an optimal learning rule should utilize those samples to learn the classification rule. One method for utilizing the explanation subgraphs is to perform data augmentation, by producing artificial training inputs with the same explanation subgraphs, and perturbed non-explanation edges. In this section, we show through a simple example that this approach may lead to worse sample complexity even compared to generic explanation-agnostic learning rules. This phenomenon is also observed in our empirical observations in the subsequent sections. As a result, while the explanations may assist in designing the learning rule, it is not necessarily helpful to use them to enlarge the training set by data augmentation without distinguishing explicitly between the artificially synthesized training elements and the original training elements. To this end, we define the data-augmentation learning rules and data-augmentation ERM (DA-ERM) as follows. 

\begin{definition}[\textbf{Explanation-Preserving Perturbation}]
\label{def:exp-preserve}
Consider a classification problem $P_G$, an explanation function $\Psi(\cdot)$, and parameter $\gamma>0$. An explanation-preserving perturbation with parameter $\gamma$ is a mapping $S^{\gamma}(\cdot)$ operating on $\overline{\mathcal{G}}$, where\footnote{$S^{\gamma}(\overline{G})$ is defined with respect to $\Psi(\cdot)$. This dependence is not made explicit in our notation to avoid clutter.}
\[{S}^{\gamma}(\overline{G})\triangleq \{\overline{G}'\Big| \Psi(\overline{G})\subseteq \overline{G}', |\mathcal{E}\Delta \mathcal{E}'|\leq \gamma|\mathcal{E}|\},\]
$\mathcal{E}$ and $\mathcal{E}'$ are the edge sets of $\overline{G}$ and $\overline{G}'$, respectively, and $\Delta$ denotes the symmetric difference.
\end{definition}
\begin{definition}[\textbf{Data-Augmentation Learning Rule}]
\label{def:EA_LR}
Given an explainer $\Psi(\cdot)$, a hypothesis class $\mathcal{H}$, and a learning rule $\mathsf{L}=(L_t)_{t\in \mathbb{N}}$,
a data-augmentation learning rule $\mathsf{L}_{exp}=(L_{exp,t})_{t\in \mathbb{N}}$ consists of a family of mappings
\begin{align*}
    L_{exp,t}:\mathcal{T}_t\mapsto \widetilde{f}(\cdot),
\end{align*}
where,
\begin{align}
\label{eq:f_aug}
    \widetilde{f}(\overline{G})\!=\!
    \begin{cases}
        Y_{exp}\quad & \exists i\in [t]\!:\! \Psi(\overline{G}_i)\subseteq \overline{G},\\
        f(\overline{G}) & \text{Otherwise}
    \end{cases},
\end{align}
$Y_{exp}$ is chosen randomly and uniformly from the set $\{Y_i|  \Psi(\overline{G}_i)\subseteq \overline{G}, i\in [t]\}$,
$f(\cdot) \triangleq L_t(\mathcal{T}_{aug,t})$,
and we have defined the augmented training set as
\begin{align*}
   \mathcal{T}_{aug,t}\triangleq\mathcal{T}_t\cup \Big(
   \bigcup_{(\overline{G},Y)\in \mathcal{T}_t} \{ (\overline{G}',Y)| \overline{G'}\in S^{\gamma}(\overline{G})\}
  \Big).
\end{align*}
The learning rule is a DA-ERM learning rule if $\mathsf{L}= \mathsf{L}_{ERM}$.
\end{definition}
For $\gamma=0$, DA-ERM is the same as EA-ERM and there is no data augmentation. The following example shows that in general, for $\gamma>0$, DA-ERM may have worse sample complexity than the explanation-agnostic ERM. 
\begin{Example}
\label{ex:DAERM}
    Consider the hypothesis class $\mathcal{H}$ which consists of all classifiers that classify their input only based on the number of edges in the graph. That is, 
    \[\mathcal{H}=\{f(\cdot)| 
   \forall \overline{G},\overline{G}': |\overline{G}|=|\overline{G}'|\rightarrow f(\overline{G})=f(\overline{G}')\}.\]
   % for any $h\in \mathcal{H}$, $h(\overline{G})= y_{|\overline{G}|}$, where $|\overline{G}|$ denotes the number of edges in $\overline{G}$, and $y_{|\overline{G}|}\in\mathcal{Y}$. 
   Furthermore, let us consider the following binary classification problem. Let $P_Y(\cdot)$ be a binary symmetric distribution, i.e., $P_Y(0)=P_Y(1)=\frac{1}{2}$. Let the graphs associated with label 0 consist of the collection 
    \[\mathcal{B}_0\!=\!\{\overline{G}\big| |\overline{G}|=n, \exists i\in [n]\!: \!C_i\!\subseteq \!\overline{G} \text{ and } \nexists j\subseteq [n]\!:\! \!D_j\!\subseteq \!\overline{G}\!\},\]
    where $n>10$ is a fixed number, $C_i, i\in [n]$ denotes a cycle of size $i$, and $D_i$ denotes a star of size $i$, where a star is a subgraph where all vertices are connected to a specific vertex called the center, and there are no edges between the rest of the vertices.  
    Thus, $\mathcal{B}_0$ consists of all graphs with exactly $n$ edges that contain at least one cycle but no stars. Similarly, let the graphs associated with label 1 be given by  
    \begin{align*}\mathcal{B}_0&\!=\!\{\overline{G}\big| |\overline{G}|=n+1, \nexists i\in [n+1]\!: \!C_i\subseteq \overline{G} 
    \\&\qquad \qquad \qquad \text{ and } \exists j\in [n+1]\!:\! \!D_j\!\subseteq \!\overline{G}\!\}.
    \end{align*}
    That is, $\mathcal{B}_1$ consists of all graphs with exactly $n+1$ edges that do not contain a cycle but contain a star. Let $\gamma= \frac{1}{n}$, and define \[\Psi(\overline{G})\triangleq 
    \begin{cases}
        C_i\qquad &\text{ if } \exists i: C_i \subseteq \overline{G},
        \\D_i \qquad &\text{ if }  \exists i: D_i \subseteq \overline{G},
    \end{cases}.
    \]
    Clearly $\zeta=0$ in this case. 
    Then, it is straightforward to see that ERM and EA-ERM both achieve zero error after observing at least one sample per label since all graphs of size $n$ have label $0$ and all graphs of size $n+1$ have label 1, and the hypothesis class decides based only on the number of edges. On the other hand,
    for DA-ERM to achieve zero error it needs to observes all possible explanation outputs, as it cannot distinguish between the augmented elements of $\mathcal{B}_0$ and the original elements of $\mathcal{B}_1$ and vice versa since they may have the same number of edges. Thus, data augmentation yields infinite sample complexity, whereas ERM and EA-ERM have sample complexity equal to two.
\end{Example}
The issue illustrated in the previous example appears to  be a fundamental issue. To explain further, note that DA-ERM empirically minimizes the risk over the augmented dataset. If the elements of the augmented dataset are in-distribution with respect to $P_G$, this also guarantees that the risk is minimized with respect to the original dataset, hence achieving similar performance as that of EA-ERM. However, if the elements of the augmented dataset are out-of-distribution with respect to $P_G$, then it may be the case that the output of DA-ERM performs well on the out-of-distribution elements, but has high error on the in-distribution elements (which are dominated by the out-of-distribution elements). Hence, DA-ERM may achieve high error probability on the original data distribution. This is exactly the phenomenon that is observed in the previous example. 
 We show this phenomenon empirically and further explain it in our empirical evaluations in the subsequent sections.

\section{Explanation-Assisted GNN Architectures}
\label{sec:learn}
\begin{figure}[!t]
\centering
\includegraphics[width=0.47\textwidth]{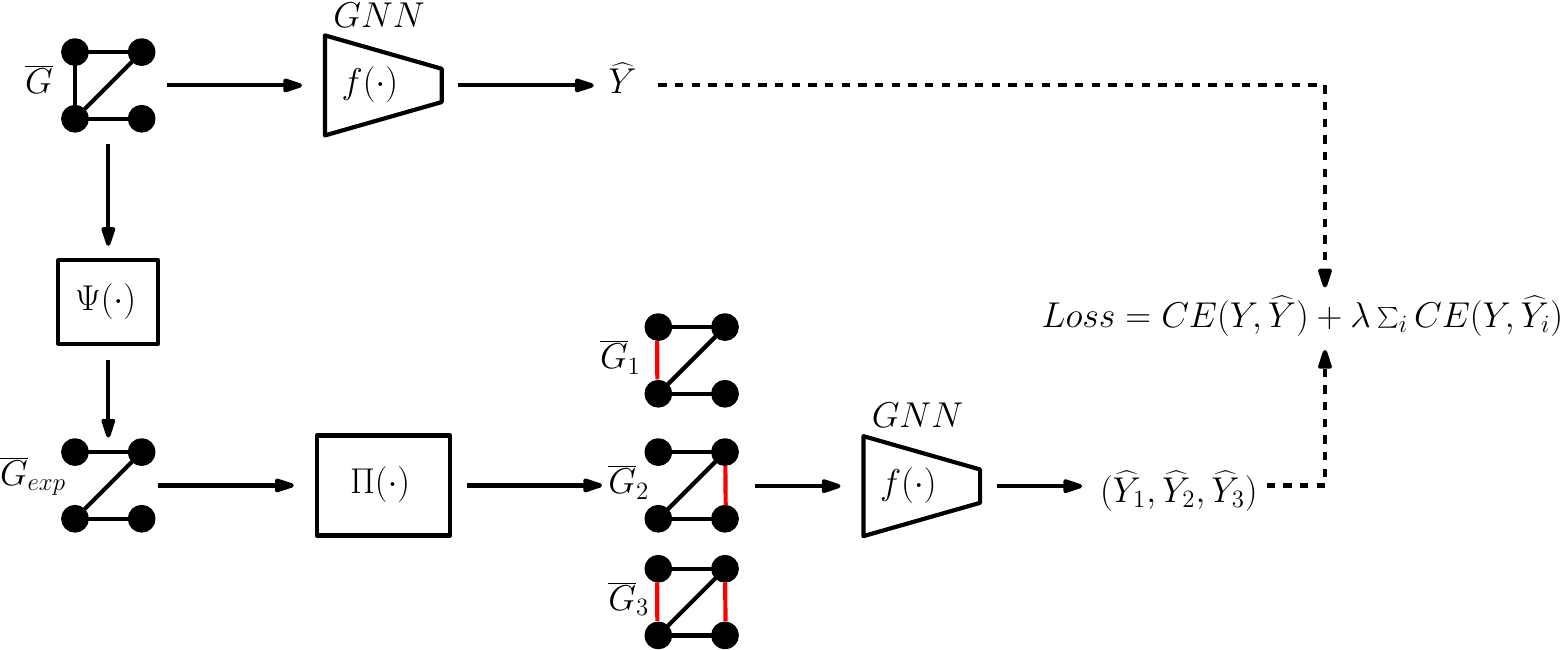}\vspace{-0.2cm}
\caption{Given training sample $(\overline{G},Y)$, the explanation subgraph 
$\overline{G}_{exp}= \Psi(\overline{G})$ is produced. Then, $\Pi(\overline{G}_{exp})$ produces the explanation-preserving perturbations $\overline{G}_i$. $\widehat{Y}$ and $\widehat{Y}_i$ are produced by passing the original and perturbed graphs through $f(\cdot)$, respectively. The loss is defined as a weighted sum of the classification loss of original and perturbed graphs. 
}
\label{fig:framework}
\end{figure}
In Section \ref{sec:PAC}, we showed that the sample complexity for EA-ERM can be arbitrarily smaller than that of standard explanation-agnostic learning methods. However, EA-ERM faces two major issues: i) empirical risk minimization across the hypothesis class is impractical for real-world applications; ii) the EA-ERM approach requires an exhaustive search over all explanation subgraphs in the training set to verify if any are a subgraph of the input. The first issue can be addressed by developing efficient learning algorithms that approximate the risk-minimizing classifier similar to traditional learning approaches. However, the second problem is intrinsic to the design of EA-ERM.

An alternative solution, explored in Section \ref{sec:aug}, is the data augmentation method. This method enlarges the training set by generating new samples through explanation-preserving perturbations of existing training samples. However, we showed through an example that if the learning algorithm does not distinguish between these artificially generated augmented samples and the original training samples, it could lead to worse performance even compared to conventional explanation-agnostic learning methods.  Building on the theoretical observations of previous sections, in this section, we introduce a practically implementable explanation-assisted GNN architecture and training procedure. 
As shown in Figure \ref{fig:framework}, given a labeled training sample $(\overline{G},Y)$ and explanation function $\Psi(\cdot)$, we first compute an explanation subgraph $\overline{G}_{exp}=\Psi(\overline{G})$. Then, we use an explanation-preserving, non-parametric perturbation operator $\Pi(\cdot)$
%\DL{Marked, $\Pi$ is a non-parametric function.  As shown in Algorithm 2.}
to produce perturbations $\overline{G}_i$ of the original input graph $\overline{G}$, such that  $\overline{G}_{exp}\subseteq \overline{G}_i$. Then, $\overline{G}$ and $\overline{G}_i$ are passed through the GNN $f(\cdot)$ to produce the output labels $\widehat{Y}$ and $\widehat{Y}_i$, respectively. The loss is defined as:
\begin{align}\label{eq:loss}
    Loss= CE(Y,\widehat{Y})+\lambda \sum_{i}CE(Y,\widehat{Y}_i). 
\end{align}
As shown in Section \ref{sec:aug}, if the perturbed graphs are out-of-distribution with respect to the input graph distribution, then the performance may be worse than explanation-agnostic methods. This is addressed in two ways in our training procedure.  First, we follow an existing work to implement the perturbation function $\Pi(\cdot)$ which randomly removes a small number of non-explanation edges~\cite{zheng2023robust} (Algorithm \ref{alg:perturb}).  As shown in previous studies, this method is effective in generating in-distribution graphs. Second, to further alleviate the negative effects of out-of-distributed augmentations, we choose the hyperparameter $\lambda$ (in Eq. \ref{eq:loss}) small enough, so that the loss on the (potentially out-of-distribution) augmented data does not dominate the loss on the original data.

\begin{algorithm}[!t]
    \centering
    \caption{Explanation-Assisted Training Algorithm}
    \label{alg:training}
\begin{footnotesize}
    \begin{algorithmic}[1]
        \STATE {\bfseries Input:} Training set $\mathcal{T}$,  %GNN model $f$, Explainer $\Psi$, 
        balancing coefficient $\lambda$, GNN pre-train epoch $e_w$, train epoch $e_s$, sampling number $M$
        \STATE {\bfseries Output:}  Trained model $f$ %, Trained explainer $\Psi(\cdot)$
        \STATE Initiate $f$, $\Psi$, $j = 0$ 
        \FOR{$ j \leq e_w $} 
            \STATE Update $f$ via $\mathbb{E}_{\mathcal{T}}(CE(Y,f(\overline{G})))$
            \STATE $j = j+1$
        \ENDFOR
        % \FOR{}
        % \STATE Train the explainer $\Psi(\cdot)$
                    \STATE Initiate empty set $\mathcal{T'}$
        \FOR{ each $(\overline{G},Y) \in \mathcal{T}$}
                \STATE $\overline{G}_{exp} = \Psi(\overline{G})$
                \FOR {$m$ in $[1,2,...M]$}
                    \STATE $\mathcal{T'} = \mathcal{T'} +\{(\Pi(\overline{G}_{exp}),Y)\}$.
                \ENDFOR
        \ENDFOR    
        \STATE Initialize $f$, $j = 0$
        \FOR{$ j \leq e_s $} 
            \STATE  Train  $f$ with $\mathbb{E}_{\mathcal{T}}(CE(Y,f(\overline{G})) + \lambda \mathbb{E}_{\{\mathcal{T'}\}}CE(Y,f(\overline{G}))) $ 
            \STATE $j = j+1$
        \ENDFOR  
    \end{algorithmic}
\end{footnotesize}
\end{algorithm}

It should be noted that the ground-truth explanation $\Psi(\overline{G})$ may not be available beforehand in real-world applications. In such scenarios, we pre-train the graph classifier $f(\cdot)$ and $\Psi(\cdot)$. This two-step training procedure is described in Algorithm \ref{alg:training}.
The proposed method is a general framework that can be employed for training various GNN architectures and explainers, such as GIN~\cite{xu2018powerful}, PNA~\cite{corso2020principal}, GNNExplaier~\cite{ying2019gnnexplainer} and PGExplainer~\cite{luo2020parameterized}.

\begin{algorithm}
    \centering
    \caption{In-distributed explanation-preserving perturbation function $\Pi(\cdot)$}
    \label{alg:perturb}
\begin{footnotesize}
    \begin{algorithmic}[1]
        \STATE {\bfseries Input:} a graph $\overline{G}$, explainer $\Psi(\cdot)$, hyper-parameter $\alpha_1$. 
        % \STATE {\bfseries Output:} binary Gumbel softmax.
        \STATE $\overline{G}^c = \overline{G}-\Psi(\overline{G})$ \hspace{3em} \comment{\#  Compute the non-explanation subgraph }
        \STATE $E_{\alpha_1}(\overline{G}^c) = $ sample $\alpha_1$ edges from $\overline{G}^c$
        \STATE {\bfseries Return} $E_{\alpha_1}(\overline{G}^c) + \Psi(\overline{G}) $         
    \end{algorithmic}
\end{footnotesize}
\end{algorithm}

\section{Empirical Verification}
\label{sec:sim}
This section empirically verifies our theoretical analysis and demonstrates the effectiveness of the proposed method.  A benchmark synthetic dataset, \bamo~\cite{luo2020parameterized}, and five real-world datasets, \mutag~\cite{luo2020parameterized}, {\benz}, {\fluo}, {\alk}~\cite{agarwal2023evaluating}, {\dd}~\cite{dobson2003distinguishing} and {\prot}~\cite{dobson2003distinguishing,borgwardt2005protein} are utilized in our empirical studies. We consider three representative GNN models: Graph Convolutional Network (GCN),  Graph Isomorphism Network (GIN), and  Principal Neighbourhood Aggregation (PNA)~\cite{corso2020principal}. Full experimental setups are shown in the Appendix~\ref{sec:app:fullsetup}.

\subsection{Comparison to Baseline Data Augmentations.}
\label{sec:exp:compare2baselines}
With this set of experiments, we aim to verify the effectiveness of our explanation-assisted graph learning algorithm. 

\begin{table}[!tp]
\centering
\fontsize{8}{9}\selectfont  
\setlength\tabcolsep{3pt}
\caption{Performance comparisons with  3-layer GNNs trained on 50 samples. The metric is classification accuracy. The best results are shown in bold font and the second best ones are underlined.} 
\label{tab:accuracy:gcngin3}
\setlength\tabcolsep{3.3pt}
\scalebox{0.95}{
\begin{tabular}{l|ccccccccccc}
    \toprule
    \multicolumn{1}{c|}{Dataset} &
    \multicolumn{1}{c}{\mutag} &
    \multicolumn{1}{c}{\benz} & 
    \multicolumn{1}{c}{\fluo} & 
    \multicolumn{1}{c}{\alk} &
    \multicolumn{1}{c}{\dd}&
    \multicolumn{1}{c}{\scriptsize{\prot}} \\
 \midrule[0.8pt]
&  \multicolumn{6}{c}{GCN} \\
Vanilla &  \ms{84.3}{3.2} & \ms{73.9}{5.2} & \ms{62.1}{4.3} & \ms{93.7}{3.2} & \ms{63.2}{6.6} & \ms{68.1}{6.1} \\
EI      &  \ms{85.6}{2.0} & \ms{75.3}{5.1} & \ms{59.3}{3.2} & \ms{92.2}{4.9} & \ms{63.6}{6.2} & \ms{69.6}{4.0} \\
ED      &  \ms{84.7}{3.4} & \ms{73.2}{4.1} & \ms{58.4}{3.7} & \ms{94.4}{1.8} & \ms{64.0}{6.0} & \ms{70.0}{4.0} \\   
ND      &  \ms{83.6}{3.5} & \ms{74.0}{3.8} & \ms{58.7}{3.0} & \ms{92.7}{3.4} & \ms{65.2}{4.2} & \ms{68.8}{3.3} \\
FD      &  \ms{84.7}{3.4} & \ms{75.2}{4.8} & \ms{57.6}{3.6} & \ms{93.8}{3.1} & \ms{62.5}{3.3} & \ms{68.6}{3.8} \\
Mixup   & \ms{67.4}{3.2} & \ms{53.9}{1.9} & \ms{52.5}{1.5} & \ms{64.3}{0.7} & \ms{56.0}{1.9} & \ms{60.8}{2.9}\\
  Aug$_\text{GE}$ & \underline{\ms{87.2}{1.4}} & \underline{\ms{76.2}{1.3}} & \textbf{\ms{66.6}{3.4}} & \underline{\ms{96.3}{1.3}} & \underline{\ms{66.1}{5.1}} & \underline{\ms{70.4}{5.9}}\\
 Aug$_\text{PE}$ & \textbf{\ms{87.2}{2.6}} & \textbf{\ms{76.5}{0.8}} & \underline{\ms{65.3}{5.0}} & \textbf{\ms{96.4}{1.1}} & \textbf{\ms{67.7}{4.3}} & \textbf{\ms{71.2}{6.3}} \\
 \midrule
 &  \multicolumn{6}{c}{GIN} \\
Vanilla & \ms{82.5}{3.7} & \ms{67.5}{5.9} & \ms{68.6}{5.2} & \ms{85.1}{10.3} & \ms{65.1}{4.3} & \ms{66.5}{4.0} \\
EI & \ms{82.8}{3.2} & \ms{71.6}{2.8} & \ms{66.8}{4.0} & \ms{87.5}{10.3} & \ms{64.7}{5.4} & \ms{65.5}{5.8} \\
ED & \ms{81.6}{3.7} & \ms{70.5}{4.3} & \ms{62.9}{5.1} & \ms{90.3}{6.4} & \ms{66.7}{3.8} & \ms{62.7}{5.3} \\
ND & \ms{82.2}{4.0} & \ms{71.3}{2.7} & \ms{64.9}{4.6} & \ms{88.9}{7.0} & \ms{66.7}{2.9} & \ms{65.6}{5.4} \\
FD & \ms{82.7}{2.9} & \ms{70.7}{2.8} & \ms{67.6}{5.1} & \ms{83.1}{11.7} & \ms{68.2}{4.3} & \ms{65.6}{5.0} \\
Mixup &\ms{74.5}{1.6} &\ms{59.0}{3.4} &\ms{51.6}{2.6} &\ms{65.8}{4.1} &\ms{58.6}{3.5} &\ms{62.2}{2.9}\\
Aug$_\text{GE}$ & \underline{\ms{86.0}{2.4}} & \textbf{\ms{75.4}{0.8}} & \underline{\ms{76.3}{2.1}} & \textbf{\ms{94.9}{1.1}} & \textbf{\ms{69.3}{5.2}} & \textbf{\ms{68.5}{5.9}} \\
Aug$_\text{PE}$ & \textbf{\ms{86.9}{1.8}} & \underline{\ms{75.4}{1.0}} & \textbf{\ms{76.5}{1.7}} &\underline{\ms{94.8}{1.1}} & \underline{\ms{67.4}{2.8}} & \underline{\ms{68.1}{5.5}} \\
\bottomrule
\end{tabular}
}
\end{table}

\noindent \textbf{Experimental Design.}
In Section \ref{sec:PAC}, we evaluate the sample complexity of explanation-assisted learning rules and quantify the resulting improvements over explanation-agnostic learning rules. We consider 3 GNN layers in each model. For each dataset, we randomly sample 50 labeled graphs for training and 10\% graphs for testing. Experiments with smaller training sizes and lightweight GNN models can be found in Appendix~\ref{sec:app:smalltrainingsize} and~\ref{sec:app:1layergnn}, respectively.  We compare with representative structure-oriented augmentations, Edge Inserting (EI), Edge  Dropping (ED), Node Dropping (ND), and Feature Dropping(FD)~\cite{ding2022data}. Recently, mixup operations have been introduced in the graph domain for data augmentation, such as $M$-mixup~\cite{wang2021mixup} and $G$-mixup~\cite{han2022g}. However, $M$-mixup is operating on the embedding space and cannot be fairly compared, and $G$-mixup does not apply to graphs with node type/features.  Instead, we use a normal Mixup as another baseline. For the explainer function $\Psi(\cdot)$ in our method, we consider two representative explainers, GNNExplainer~\cite{ying2019gnnexplainer} and PGExplainer~\cite{luo2020parameterized}, whose corresponding augmentations are denoted by Aug$_\text{GE}$ and Aug$_\text{PE}$, respectively.  More comprehensive results on different settings and full experimental results are shown in Appendix~\ref{sec:app:extraexp}.

\stitle{Experimental Results.} From Tables~\ref{tab:accuracy:gcngin3} and \ref{tab:accuracy:pna3} (in the Appendix), we have the following observations. First, our explanation-assisted learning methods consistently outperform the vanilla GNN models as well as the ones trained with structure-oriented augmentations by large margins.  Utilizing the GCN as the backbone, our methods, Aug$_\text{GE}$ and Aug$_\text{PE}$, exhibit significant enhancements in classification accuracy—2.40\% and 2.75\%, on average—when compared to the best-performing baselines across six datasets. With GIN, the improvements are 5.04\% and 4.69\%, respectively. Secondly, we observe that traditional structure-based augmentation methods yield comparatively less effectiveness. For example, in the {\fluo} dataset, all baseline augmentation methods achieve negative effects, while our methods can still beat the backbone significantly.

\subsection{Effects of Augmentation Distribution }
\label{sec:exp:augmentationdistribution}
In Section \ref{sec:aug}, we investigated explanation-assisted data augmentation methods, and argued that performance improvements are contingent on in-distribution generation of augmented data. In this section, we analyze the effects of augmentation distributions on the model accuracy. With this set of experiments on a synthetic dataset and a real-world dataset, we aim to explore two questions: (RQ1) Can in-distribution augmentations lead to better data efficiency in graph learning? (RQ2) What are the effects of out-of-distribution augmentations on graph learning?

\begin{figure}[h]
\vspace{-0.3cm}
  \centering
    \subfigure[\bamo]{\includegraphics[width=0.25\textwidth]{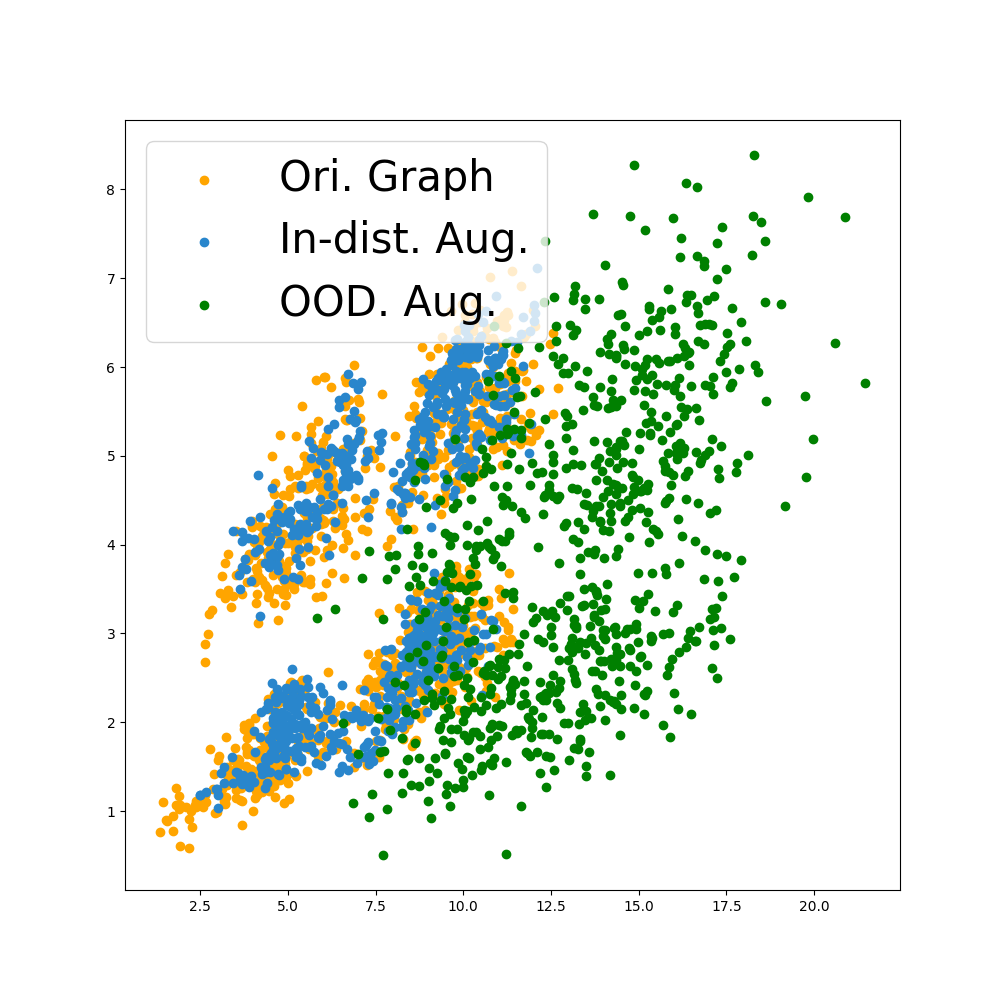} \hspace{-1.5em}\label{fig:exp1:distribution:bamo}}    
    \subfigure[\benz]{\includegraphics[width=0.25\textwidth]{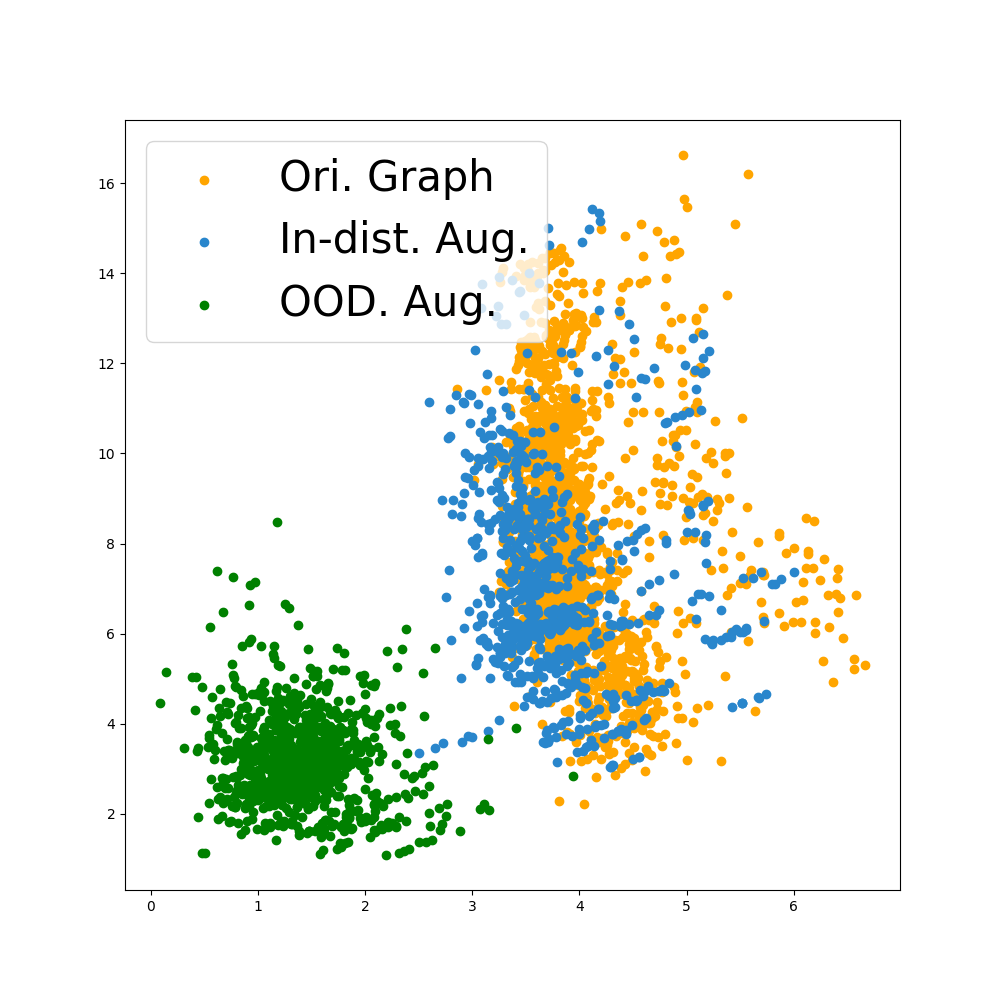}\label{fig:exp1:distribution:benz}} \vspace{-0.2cm}
    \caption{The visualization of original graphs, in-distributed and OOD augmentations in dataset \bamo~ and \benz. }
\vspace{-0.2cm}
    \label{fig:exp1:distribution}
\vspace{-0.3cm}
\end{figure}

To evaluate the data efficiency of graph learning methods, we vary the number of training samples in the range $[4,8,20,40,100,300,500,700]$. We sufficiently train GNN models with three settings: 1) training with the vanilla training samples, 2) training with in-distribution explanation-preserving augmentation, and 3) training with out-of-distribution explanation-preserving augmentation.  For setting 2, we use the proposed augmentation method on the ground truth explanations. For setting 3, to generate OOD augmentations, we randomly add 100\% edges from the BA graph for each instance on the {\bamo} dataset. On the {\benz} dataset, we randomly remove 30\% edges from the non-explanation subgraphs.  Visualization results on three sets of graphs are shown in Figure~\ref{fig:exp1:distribution}, which shows that our methods are able to generate both in-distributed and out-of-distributed augmentations for further analysis.

\begin{figure}[h]
  \centering
    \subfigure[GCN on \bamo]{\includegraphics[width=1.50in]{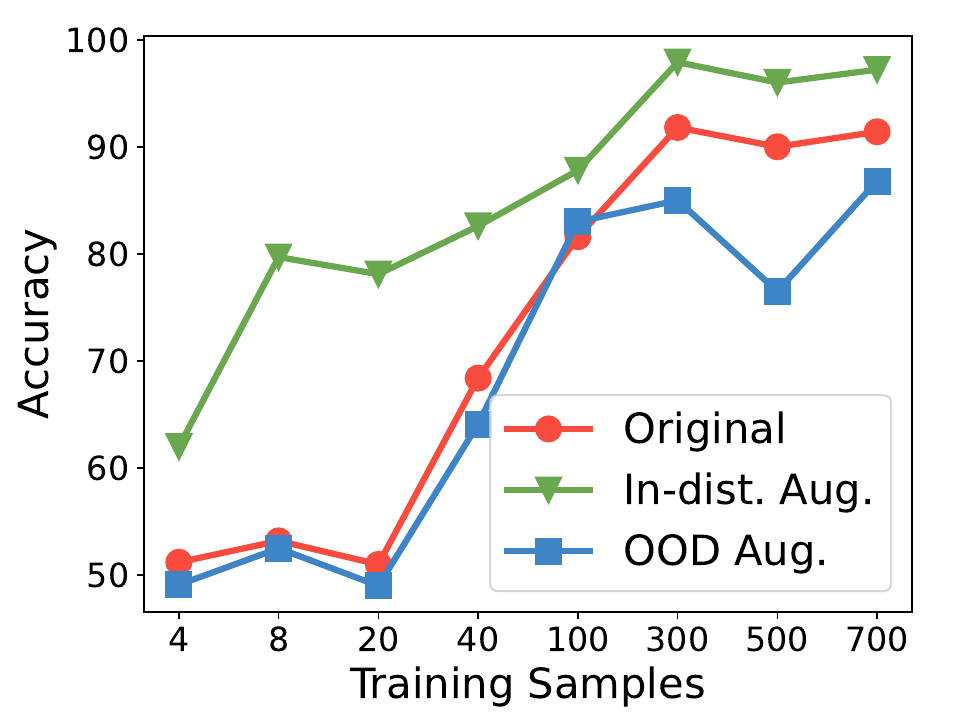} \hspace{-0.1em}\label{fig:exp:accuracy:gcnbamo}  }
    \subfigure[GCN on \benz]{\includegraphics[width=1.50in]{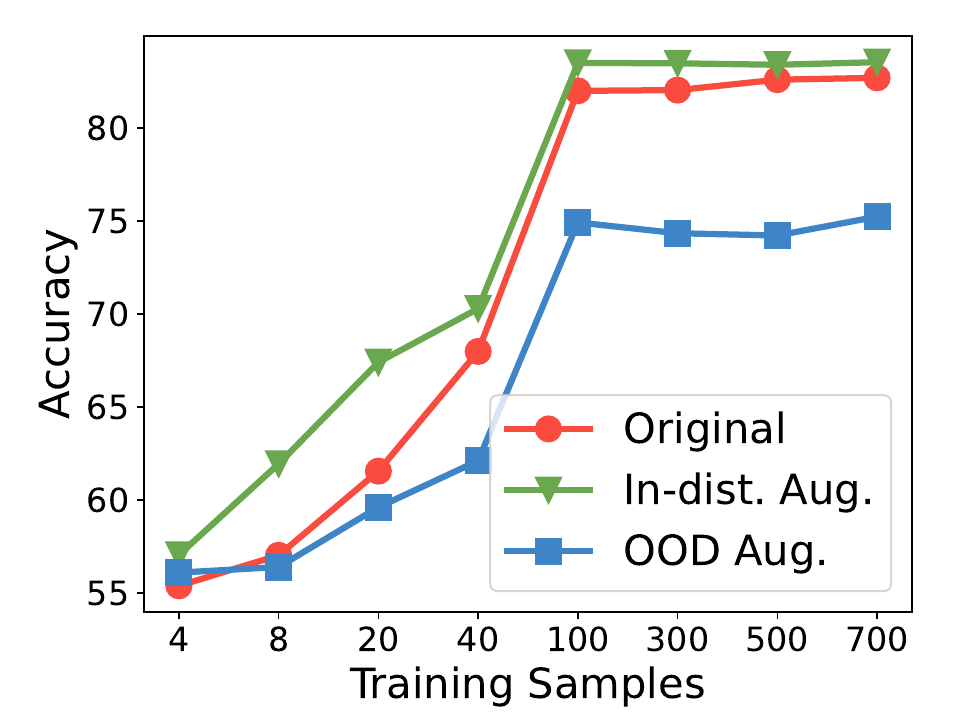} \hspace{-0.1em}\label{fig:exp:accuracy:ginbamo}}\\
    \subfigure[GIN on \bamo]{\includegraphics[width=1.50in]{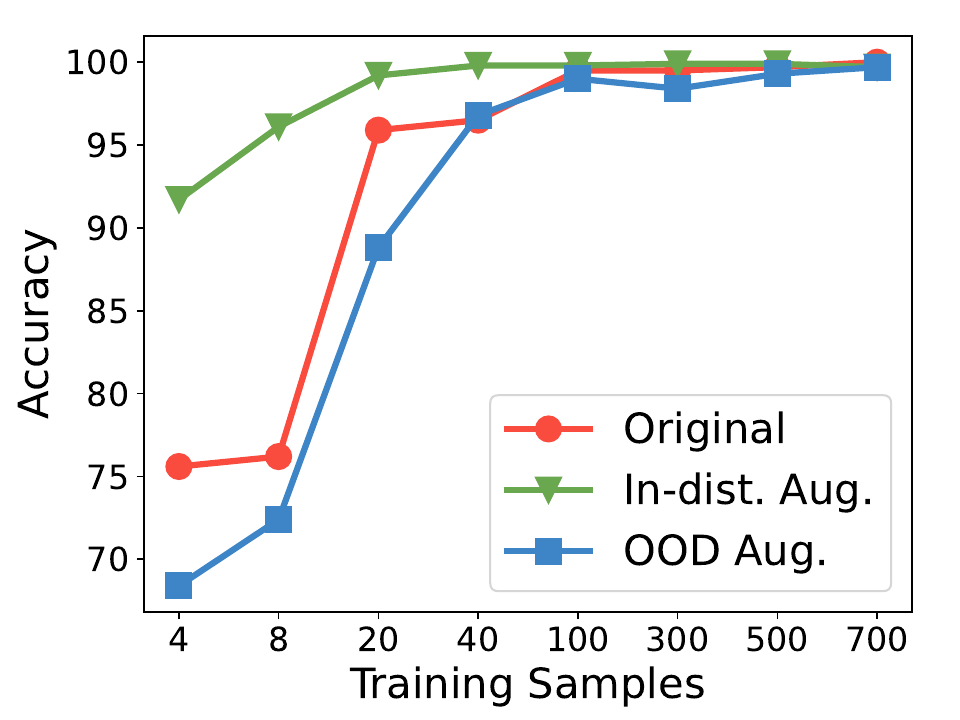} \hspace{-0.1em}\label{fig:exp:accuracy:gcnbahouse}} 
    \subfigure[GIN on \benz]{\includegraphics[width=1.50in]{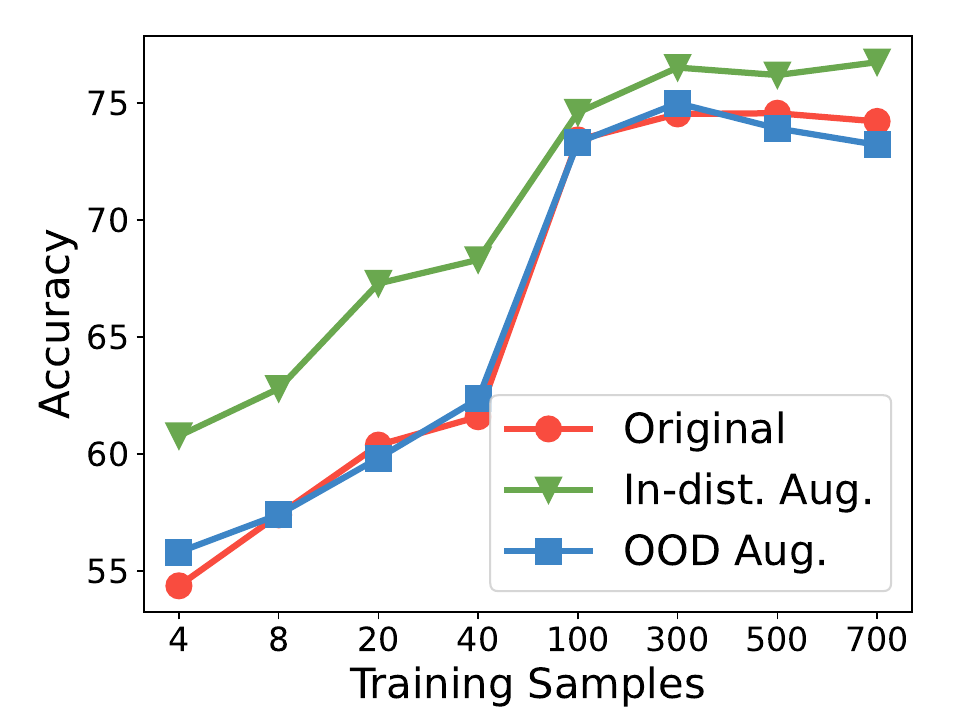} \label{fig:exp:accuracy:ginbahouse}}
    \caption{Effects of in-distributed and OOD augmentations on the accuracy of GCN and GIN on {\bamo} and {\benz} datasets.}
    \label{fig:exp:accuracy}
        \vspace{-0.3cm}
\end{figure}

We answer our research questions with accuracy performances in Figure~\ref{fig:exp:accuracy}. From these figures, we have the following observations. First, in-distribution augmentations significantly and consistently improve the data efficiency of both GCN and GIN in two datasets. For example, with explanation-preserving augmentations, GIN can achieve over 90\% accuracy with only 4 samples in the synthetic dataset, while the performance of GIN trained with original datasets is around 75\%.  Second, OOD augmentations fail to improve data efficiency in most cases. Moreover, for GCN on {\benz}, the OOD augmentation worsens the performance, which is aligned with our theoretical analysis.

\section{Additional Related Work} 
\label{sec:relatedwork}
% \subsection{Graph Neural Networks} 
\textbf{Explainable Graph Neural Networks.} 
%Understanding how the GNN models work and providing explanations is an important topic in recent years.
Prior works~\cite{ying2019gnnexplainer,luo2020parameterized,yuan2020xgnn,yuan2022explainability,yuan2021explainability, lin2021generative, wang2022gnninterpreter,miao2023interpretable,fang2023cooperative,xie2022task,ma2022clear} have explored various methods to provide interpretability in GNNs. Traditional methods obtain explanation by gradient, such as SA~\cite{sa_baldassarre2019explainability}, Grad-CAM~\cite{GradCAM_pope2019explainability}. To develop model-agnostic explainers, perturbation-based methods, surrogate methods, and generation-based methods have been proposed.  In perturbation-based methods, including GNNExplainer~\cite{ying2019gnnexplainer}, PGExplainer~\cite{luo2020parameterized}, ReFine~\cite{wang2021towards}, they generate perturbation to determine which features and subgraph structure are important. Surrogate methods~\cite{vu2020pgm,duval2021graphsvx} use a surrogate model to approximate the local prediction and use this surrogate model to generate explanations. Generation-based methods~\cite{yuan2020xgnn,shan2021reinforcement,wang2022gnninterpreter} adopt generative models to derive instance-level or model-level explanations.

\textbf{Data Augmentation.}
Data augmentation is a widely used technology in self-supervised learning~\cite{you2020graph,zhu2020deep}. A large class of graph augmentation methods can be categorized as rule-based~\cite{wang2021mixup,rong2019dropedge,gasteiger2019diffusion,zhao2022learning} and learning-based methods~\cite{zhao2021data,wu2022discovering,zhao2022autogda}. Rule-based methods include NodeDrop~\cite{rong2019dropedge}, EdgeDrop~\cite{feng2020graph}, MessageDrop~\cite{fang2023dropmessage}, which randomly drop some features and information from the original graph. GraphCrop~\cite{wang2020graphcrop} and MoCL~\cite{sun2021mocl} randomly crop and substitute the graphs. 
Learning-based methods use graph neural networks to learn which edge is important, for example, ProGNN~\cite{jin2020graph} learns a structural graph from a poisoned graph. GraphAug~\cite{luo2022automated} proposed a reinforcement learning method to produce the label-invariant augmentations. Besides, in Half-Hop\cite{azabou2023half}, authors proposed a novel graph augmentations by inserting a slow node. In \cite{liu2022local}, a local augmentation is proposed by learning the conditional distribution of the node under its neighbors.

\section{Conclusion}
Two approaches for leveraging subgraph explanations in graph learning were considered. First, explanation-assisted learning rules were considered, where the explanation subgraphs are directly fed to the learning mechanism along with labeled training samples. The sample complexity was characterized and was shown to be arbitrarily smaller than the explanation-agnostic sample complexity. Next, explanation-assisted data augmentation was considered, followed by a generic learning rule applied to the augmented dataset. It was shown both theoretically and empirically that this may sometimes lead to better and sometimes to worse performance in terms of sample complexity, where gains are contingent on producing in-distribution augmented samples. 

\bibliography{ref}
\bibliographystyle{icml2024}

\clearpage

%%%%%%%%%%%%%%%%%%%%%%%%%%%%%%%%%%%%%%%%%%%%%%%%%%%%%%%%%%%%%%%%%%%%%%%%%%%%%%%
%%%%%%%%%%%%%%%%%%%%%%%%%%%%%%%%%%%%%%%%%%%%%%%%%%%%%%%%%%%%%%%%%%%%%%%%%%%%%%%
% APPENDIX
%%%%%%%%%%%%%%%%%%%%%%%%%%%%%%%%%%%%%%%%%%%%%%%%%%%%%%%%%%%%%%%%%%%%%%%%%%%%%%%
%%%%%%%%%%%%%%%%%%%%%%%%%%%%%%%%%%%%%%%%%%%%%%%%%%%%%%%%%%%%%%%%%%%%%%%%%%%%%%%
\newpage
\appendix
\onecolumn

\section{Proofs}
\subsection{Proof of Proposition \ref{Prop:1}}
\label{App:Prop:1}
Since the optimal Bayes error of predicting $Y$ from $\overline{G}$ is less than $\overline{\epsilon}$, from Fano's inequality, we must have:
\begin{align}
\label{App:1:Eq:1}
  H(Y|\overline{G},\Psi(\overline{G}))= H(Y|\overline{G})\leq h_b(\overline{\epsilon}),
\end{align}
where $h_b(\cdot)$ denotes the binary entropy function. On the other hand, from $(\kappa,s)$-explainability assumption and Condition 1, we have:
\begin{align}
\label{App:1:Eq:2}
I(Y;\overline{G}|\Psi(\overline{G}))= H(Y|\Psi(\overline{G}))-H(Y|\overline{G}) \leq \overline{\kappa}.
\end{align}
Combining \eqref{App:1:Eq:1} and \eqref{App:1:Eq:2}, we conclude that:
\begin{align*}
    H(Y|\Psi(\overline{G}))\leq \overline{\kappa}+h_b(\overline{\epsilon}). 
\end{align*}
Let $\epsilon'\triangleq \overline{\kappa}+h_b(\overline{\epsilon})$. From reverse Fano's inequality \citep{tebbe1968uncertainty,kovalevsky1968problem,sakai2017sharp}, it follows that the Bayes error of the  estimation of $Y$ based on  $\Psi(\overline{G})$ goes to zero as $\epsilon'\to 0$. So, there exist $\overline{\kappa},\overline{\epsilon}$ small enough and a classification function $f_{exp}$ operating on $\Psi(\overline{G})$ such that $P(f_{exp}(\Psi(\overline{G}))\neq Y_{\overline{G}})\leq \frac{\zeta}{2}$. %From Condition 1, we conclude that for any $\overline{G}'\in \mathcal{S}^{\gamma}(\overline{G})$, we have $\Psi(\overline{G})=\Psi(\overline{G}')$.
Consequently, 
\begin{align*}
&\sum_{g_{exp}}P(\Psi(\overline{G})=g_{exp}) P(Y_{\overline{G}'}\neq Y_{\overline{G}''}|\Psi(\overline{G}')=\Psi(\overline{G}'') =g_{exp})
\\& \leq  \sum_{g_{exp}}P(\Psi(\overline{G})=g_{exp}) P(Y_{\overline{G}'}\neq f_{exp}(g_{exp}) \text{ or 
 }Y_{\overline{G}''}\neq f_{exp}(g_{exp})|\Psi(\overline{G}')=\Psi(\overline{G}'') =g_{exp})
 \\& \stackrel{(a)}{\leq} \sum_{g_{exp}}P(\Psi(\overline{G})=g_{exp}) P(Y_{\overline{G}'}\neq f_{exp}(g_{exp})|\Psi(\overline{G}')=\Psi(\overline{G}'') =g_{exp})
 \\& +
 \sum_{g_{exp}}P(\Psi(\overline{G})=g_{exp}) P(Y_{\overline{G}''}\neq f_{exp}(g_{exp})|\Psi(\overline{G}')=\Psi(\overline{G}'') =g_{exp})
 \\&\stackrel{(b)}{=} \sum_{g_{exp}}P(\Psi(\overline{G}')=g_{exp}) P(Y_{\overline{G}'}\neq f_{exp}(g_{exp})|\Psi(\overline{G}')=g_{exp})
 \\& +
 \sum_{g_{exp}}P(\Psi(\overline{G}'')=g_{exp}) P(Y_{\overline{G}''}\neq f_{exp}(g_{exp})|\Psi(\overline{G}'') =g_{exp})
 \\& = 2P(Y_{\overline{G}}\neq f_{exp}(\Psi(\overline{G})))\leq \zeta,
\end{align*}
where in (a) we have used the union bound, and in (b) we have used the fact that $\overline{G}'$ and $\overline{G}''$ are independent of each other. 

\qed

\subsection{Proof of Theorem \ref{th:1}}
\label{App:th:1}
\begin{proof}
\textit{Proof of the upper-bound on the sample complexity:} We prove that the EA-ERM learning rule (Definition \ref{def:EAERM}) achieves the sample complexity in the theorem statement. The proof builds upon the techniques developed for evaluating the sample complexity under transformation invariances in \cite{shao2022theory}. 
\\ Fix $m\in \mathbb{N}$.  Let $d= VC_{EA}(\mathcal{H},\Psi,\gamma)$. Consider two sets $\mathcal{T}$ and $\mathcal{T}'$ of $m$ independently generated labeled random graphs generated according to $P_G$. Let us denote the event \[\mathcal{E}_\mathcal{T,\epsilon}\triangleq \{\exists f\in \mathcal{H}: err_{P_G}(\widetilde{f})\geq err_{\mathcal{T}}(f)+\epsilon\},\] where $err_{\mathcal{T}}(\cdot)$ denotes the empirical error over the set $\mathcal{T}$, $err_{P_G}$ denotes the statistical error with respect to $P_G$, and $\widetilde{f}(\cdot)$ is defined based on $f(\cdot)$ and $\mathcal{T}$ as in \eqref{eq:f_aug}. In order to derive an upper-bound on the sample complexity, we first provide sufficient conditions on $m$ under which $P(\mathcal{E}_{\mathcal{T},\epsilon})\leq \frac{\delta}{2}$. To this end, let us define \[\mathcal{E}_{\mathcal{T},\mathcal{T}',\epsilon}=\{\exists f \in \mathcal{H}: err_{\mathcal{T}'}(\widetilde{f})\geq err_{\mathcal{T}}(f)+\frac{\epsilon}{2}\}.\]

Note that:
\begin{align}
   &\label{eq:App1:1} P(\mathcal{E}_{\mathcal{T},\mathcal{T}',\epsilon})\geq P( \mathcal{E}_{\mathcal{T},\epsilon} ,  \mathcal{E}_{\mathcal{T},\mathcal{T}',\epsilon})= P( \mathcal{E}_{\mathcal{T},\epsilon})P(\mathcal{E}_{\mathcal{T},\mathcal{T}',\epsilon}| \mathcal{E}_{\mathcal{T},\epsilon}),
   \end{align}
   Consequently, to derive an upper-bound on $P( \mathcal{E}_{\mathcal{T},\epsilon})$ it suffices to derive a lower-bound on $P(\mathcal{E}_{\mathcal{T},\mathcal{T}',\epsilon}| \mathcal{E}_{\mathcal{T},\epsilon})$ and an upper-bound on $P(\mathcal{E}_{\mathcal{T},\mathcal{T}',\epsilon})$. We first derive a lower-bound on $P(\mathcal{E}_{\mathcal{T},\mathcal{T}',\epsilon}| \mathcal{E}_{\mathcal{T},\epsilon})$. Note that:
   \begin{align}
   &\nonumber P(\mathcal{E}_{\mathcal{T},\mathcal{T}',\epsilon}| \mathcal{E}_{\mathcal{T},\epsilon})=
   P(\exists f'\in \mathcal{H}: err_{\mathcal{T}'}(\widetilde{f'})\geq err_{\mathcal{T}}(f')+\frac{\epsilon}{2}\big| \exists f\in \mathcal{H}: err_{P_G}(\widetilde{f})\geq err_{\mathcal{T}}(f)+\epsilon)
   \\&\nonumber\geq P( err_{\mathcal{T}'}(\widetilde{f})\geq err_{\mathcal{T}}(f)+\frac{\epsilon}{2}\big| err_{P_G}(\widetilde{f})\geq err_{\mathcal{T}}(f)+\epsilon)
   \\&\nonumber = \sum_{(\overline{g}_i,y_i), i\in [m]}P(\mathcal{T}=\{(\overline{g}_i,y_i)| i\in [m]\})P( err_{\mathcal{T}'}(\widetilde{f})\geq err_{\mathcal{T}}(f)+\frac{\epsilon}{2}\big| err_{P_G}(\widetilde{f})\geq err_{\mathcal{T}}(f)+\epsilon, \mathcal{T}=\{(\overline{g}_i,y_i)| i\in [m]\})
\end{align}
For a given realization of the training set $\mathcal{T}=\{(\overline{g}_i,y_i), i \in [m]\}$, let $err_{\mathcal{T}}(f)+\epsilon$ be denoted by the (constant) variable $c_{\mathcal{T}}$. Then, we have:
\begin{align*}
     &P(\mathcal{E}_{\mathcal{T},\mathcal{T}',\epsilon}| \mathcal{E}_{\mathcal{T},\epsilon})=
     \sum_{(\overline{g}_i,y_i), i\in [m]}P\left(\mathcal{T}=\{(\overline{g}_i,y_i)| i\in [m]\})P( err_{\mathcal{T}'}(\widetilde{f})-c_{\mathcal{T}}\geq -\frac{\epsilon}{2}\Big| \mathbb{E}(err_{\mathcal{T}'}(\widetilde{f}))\geq c_{\mathcal{T}}\right)
\end{align*}
where we have used the fact that $\mathcal{T}'$ is a collection of independent and identically distributed (IID) samples to conclude that $err_{P_G}(\widetilde{f})= \mathbb{E}(err_{\mathcal{T}'}(\widetilde{f}))$. Consequently, 
\begin{align*}
    P(\mathcal{E}_{\mathcal{T},\mathcal{T}',\epsilon}
    \big | \mathcal{E}_{\mathcal{T},\epsilon})\geq 
    P( err_{\mathcal{T}'}(\widetilde{f})- \mathbb{E}(err_{\mathcal{T}'}(\widetilde{f}))\geq -\frac{\epsilon}{2}\big),
\end{align*}
where we have dropped the condition $ \mathbb{E}(err_{\mathcal{T}'}(\widetilde{f}))\geq c_{\mathcal{T}}$ since $err_{\mathcal{T}'}(\widetilde{f})- \mathbb{E}(err_{\mathcal{T}'}(\widetilde{f}))$ is independent of $\mathcal{T}$, and used the fact that $\sum_{(\overline{g}_i,y_i), i\in [m]}P(\mathcal{T}=\{(\overline{g}_i,y_i)| i\in [m]\})=1$.
Note that $err_{\mathcal{T}'}(\widetilde{f})= \frac{1}{m} \sum_{(\overline{G}_i,Y_i)\in \mathcal{T}'}\mathds{1}(\widetilde{f}(\overline{G}_i)\neq Y_i)$ and $ \mathbb{E}(err_{\mathcal{T}'}(\widetilde{f}))= P(\overline{f}(G_i)\neq Y_i), i\in [m]$, where we have used the linearity of expectation. So, 
\begin{align*}
   P(\mathcal{E}_{\mathcal{T},\mathcal{T}',\epsilon}| \mathcal{E}_{\mathcal{T},,\epsilon})\geq 
   P( \sum_{(\overline{G}_i,Y_i)\in \mathcal{T}'}(\mathds{1}(\widetilde{f}(\overline{G}_i)\neq Y_i)-  P(\overline{f}(G_i)\neq Y_i))\geq \frac{-m\epsilon}{2})\geq 1-e^{-\frac{2m^2\epsilon^2}{m}}= 1-e^{-2m\epsilon^2},
\end{align*}
where we have used the Hoeffding's inequality and the fact that $\mathds{1}(\widetilde{f}(\overline{G}_i)\neq Y_i)\in [0,1]$. Hence, if $m\geq\frac{2}{\epsilon^2}$, then:
\begin{align}
     P(\mathcal{E}_{\mathcal{T},\mathcal{T}',\epsilon}| \mathcal{E}_{\mathcal{T},\epsilon})\geq 1-\frac{1}{e}\geq \frac{1}{2}.\label{eq:App1:2} 
\end{align}
Combining \eqref{eq:App1:1} and \eqref{eq:App1:2}, we get:
\begin{align*}
    P(\mathcal{E}_{\mathcal{T},\mathcal{T}',\epsilon})\geq \frac{1}{2}P(\mathcal{E}_{\mathcal{T},\epsilon}). 
\end{align*}
Hence, to provide an upper-bound on $P(\mathcal{E}_{\mathcal{T},\epsilon})$, it suffices to provide sufficient conditions on $m$ such that $P(\mathcal{E}_{\mathcal{T},\mathcal{T}',\epsilon})\leq \delta$. To this end, we first introduce the notion of perturbed subset. For a given set of labeled random graphs $\mathcal{U}$ and an element $(\overline{G},Y)$ in $\mathcal{U}$, let us define the perturbed set of $\overline{G}$ in $\mathcal{U}$ as $\mathcal{O}(\overline{G}|\mathcal{U})\triangleq \{(\overline{G}',Y)\in \mathcal{U}| \Psi(\overline{G})\in \overline{G}'\}$, i.e., as the set of all elements of $\mathcal{U}$ that can be produced via explanation-preserving perturbations of $\overline{G}$. Furthermore, let $n_{\mathcal{O}}(\overline{G}|\mathcal{U})\triangleq  |\mathcal{O}(\overline{G}|\mathcal{U})|$ be the number of explanation-preserving perturbations of $\overline{G}$ in $\mathcal{U}$. 

Next, let us define $\mathcal{T}''=\mathcal{T}\cup\mathcal{T}'$. 
Let $\overline{G}_1,\overline{G}_2,\cdots,\overline{G}_{2m}$ be the elements of $\mathcal{T}''$ sorted such that $n_{\mathcal{O}}(\overline{G}_i|\mathcal{T}'')\geq n_{\mathcal{O}}(\overline{G}_j|\mathcal{T}''), j\leq i$, i.e. sorted in a non-decreasing order with respect to $n_{\mathcal{O}}(\cdot|\mathcal{T}'')$ so that there are a larger or equal number of samples which are perturbations of $\overline{G}_i$ in $\mathcal{T}''$ than that of $\overline{G}_j$ for all $j\leq i$. We construct the sets $\mathcal{R}_1$ and $\mathcal{R}_2$ partitioning $\mathcal{T}''$ as follows. Initiate $\mathcal{R}_1=\mathcal{R}_2=\phi$ and $\mathcal{T}''_1=\mathcal{T}''$. 
If $n_{\mathcal{O}}(\overline{G}_1|\mathcal{T}''_1)>\log^2{m}$, we add $\mathcal{O}(\overline{G}_1|\mathcal{T}''_1)$ to $\mathcal{R}_1$ and construct $\mathcal{T}''_2= \mathcal{T}''-\mathcal{O}(\overline{G}_1|\mathcal{T}'')$. We define $\overline{G}^{(1)}=\overline{G}_1$ and call it the subset representative for $\mathcal{O}(\overline{G}_1|\mathcal{T}'')$. Next, we arrange the elements of $\mathcal{T}''_2$ in a non-decreasing order with respect to $n_{\mathcal{O}}(\cdot|\mathcal{T}''_2)$ similar to the previous step.
Let $\overline{G}^{(2)}$ denote the sample with the largest $n_{\mathcal{O}}(\cdot|\mathcal{T}''_1)$ value. If $n_{\mathcal{O}}(\overline{G}^{(2)}|\mathcal{T}''_2)\geq \log^2{m}$, its corresponding set $\mathcal{O}(\overline{G}^{(2)}|\mathcal{T}''_2)$ is added to $\mathcal{R}_1$. This process is repeated until the $\ell$th step when $n_{\mathcal{O}}(\overline{G}^{(\ell)}|\mathcal{T}_{\ell})$ is less than $\log^2{m}$. Then, we set $\mathcal{R}_2=\mathcal{T}_{\ell}$, thus partitioning $\mathcal{T}''$ into two sets. Loosely speaking, $\mathcal{R}_1$ contains the samples which have more than $\log^2{m}$ of their explanation-preserving perturbations in non-overlapping subsets of $\mathcal{T}''$, and $\mathcal{R}_2$ contains the samples which, after removing elements of $\mathcal{R}_1$, do not have more that $\log^2{m}$ of their perturbations in the other remaining samples. 

%, consisting of elements whose explanation-preserving perturbations are in $\mathcal{T}''$ more than $\log^2{m}$ and those which are not. That is, we define $\mathcal{R}_1= \{(\overline{G},Y)\big| |\mathcal{S}^{\gamma}(\overline{G})\cap \mathcal{T}''|\geq \log^2{m}\}$, and $\mathcal{R}_2=\mathcal{T}''-\mathcal{R}_1$.
%
We further define $\mathcal{S}_i= \mathcal{T}\cap \mathcal{R}_i$ and  $\mathcal{S}'_i= \mathcal{T}\cap \mathcal{R}'_i, i\in \{1,2\}$. 
For any collection $\mathcal{A}=\{(\overline{g}_i,y_i), i\in [|\mathcal{A}|]\}$ of labeled random graphs and graph classification algorithm $f'(\cdot)$, let $\overline{M}_{\mathcal{A}}(f')\triangleq \frac{1}{|\mathcal{A}|}\sum_{(\overline{g}_i,y_i)\in \mathcal{A}}\mathds{1}(f'(\overline{g}_i)\neq y_i)$ be the average number of missclassifed elements of $\mathcal{A}$ by $f'(\cdot)$. Then,
\begin{align}
    &\nonumber P(\mathcal{E}_{\mathcal{T},\mathcal{T}',\epsilon})=P(\exists f\in \mathcal{H}: err_{\mathcal{T}'}(\widetilde{f})\geq err_{\mathcal{T}}(f)+\frac{\epsilon}{2}) 
    \\&\nonumber \leq P(\exists f\in \mathcal{H}: \overline{M}_{\mathcal{S}'_1}(\widetilde{f})\geq \overline{M}_{\mathcal{S}_1}(\widetilde{f})+\frac{\epsilon}{4} \text{ or } \overline{M}_{\mathcal{S}'_2}(\widetilde{f})\geq \overline{M}_{\mathcal{S}_2}(\widetilde{f})+\frac{\epsilon}{4})
    \\& \label{eq:App1:3}
    \leq P(\exists f\in \mathcal{H}: \overline{M}_{\mathcal{S}'_1}(\widetilde{f})\geq \overline{M}_{\mathcal{S}_1}(\widetilde{f})+\frac{\epsilon}{4})
    +
    P(\exists f\in \mathcal{H}: \overline{M}_{\mathcal{S}'_2}(\widetilde{f})\geq \overline{M}_{\mathcal{S}_2}(\widetilde{f})+\frac{\epsilon}{4}).
\end{align}
We upper-bound each term in \eqref{eq:App1:3} separately. 
\\\textbf{Step 1:} Finding an upper-bound for the term $ P(\exists f\in \mathcal{H}: \overline{M}_{\mathcal{S}'_1}(\widetilde{f})\geq \overline{M}_{\mathcal{S}_1}+\frac{\epsilon}{4})$:

We first find an upper bound for $\ell$. Note that by construction
\\i) $\mathcal{R}_1= \bigcup_{i\in [\ell]}\mathcal{O}(\overline{G}^{(i)}|\mathcal{T}_i)$, 
\\ii) $n_{\mathcal{O}}(\overline{G}^{(i)}|\mathcal{T}_{i}) \geq \log^2{m}, i\in [\ell]$, 
\\iii) $\mathcal{O}(\overline{G}^{(i)}|\mathcal{T}_i)$ are disjoint, and 
\\iv) $|\mathcal{R}_1|\leq |\mathcal{T}''|=2m$. 
\\From i) and iv), we have $\bigcup_{i\in [\ell]}\mathcal{O}(\overline{G}^{(i)}|\mathcal{T}_i) \leq 2m$, and from iii), we have $\sum_{i\in [\ell]}|\mathcal{O}(\overline{G}^{(i)}|\mathcal{T}_i)|= \sum_{i\in \ell}  n_{\mathcal{O}}(\overline{G}^{(i)}|\mathcal{T}_{i}) \leq 2m$, and from ii), we conclude that $\ell\leq \frac{2m}{\log^2{m}}$.

%Next, let $\mathcal{E}_{\mathcal{O}}= \{i| \widetilde{f}(\overline{G}^{(i)})\neq Y^{(i)}, i\in [\ell]\}$, where $Y^{(i)}$ is the label of $\overline{G}^{(i)}$. That is, $\mathcal{E}_{\mathcal{O}}$ is the indices of the subset representatives which are missclassified by $\widetilde{f}$. 
Next, we bound the expected number of missclassified elements of $\mathcal{R}_1$ for which there is at least one training sample in $\mathcal{T}$ with the same explanation subgraph. To this end, let us define:
 \[
 err_{\mathcal{O}}\triangleq \frac{1}{m}\sum_{(\overline{G},Y)\in \mathcal{T}''} \mathds{1}(\widetilde{f}(\overline{G})\neq Y \land \exists (\overline{G}',Y')\in \mathcal{T}: \Psi(\overline{G}')\subseteq \overline{G}),\]
that is, $err_{\mathcal{O}}$ is the fraction of elements of $\mathcal{T}''$ which are missclassified despite the existence of at least one training sample whose explanation subgraph is a subgraph of the missclassfied graph.  Note that from Condition 1, from $\Psi(\overline{G}')\subseteq \overline{G}$ we can conclude that $\Psi(\overline{G})=\Psi(\overline{G}')$. Let $\mathcal{G}_{exp}$ be the image of $\Psi(\cdot)$,
and define 
\[err_{\mathcal{O},g_{exp}}
\triangleq  \frac{1}{m}\sum_{(\overline{G}'',Y_{\overline{G}''})\in \mathcal{T}''} \mathds{1}(\widetilde{f}(\overline{G}'')\neq Y_{\overline{G}''} \land \exists (\overline{G}',Y')\in \mathcal{T}: \Psi(\overline{G}')=\Psi(\overline{G}'')= g_{exp}), \quad g_{exp}\in \mathcal{G}_{exp}.
\]
Note that $err_{\mathcal{O}}= \sum_{g_{exp}\in \mathcal{G}_{exp}} err_{\mathcal{O},g_{exp}}$. Furthermore, \[\mathbb{E}(err_{\mathcal{O},g_{exp}})= \frac{1}{m}|\mathcal{T}''| P(\Psi(\overline{G})=g_{exp})P(Y_{\overline{G}'}\neq Y_{\overline{G}''}|\Psi(\overline{G}')=\Psi(\overline{G}'')=g_{exp}).\] Consequently,  from Proposition \ref{Prop:1}, we have 
\[\mathbb{E}(err_{\mathcal{O}})\leq \frac{1}{m}|\mathcal{T}''|
\sum_{g_{exp}}P(\Psi(\overline{G})=g_{exp})P(Y_{\overline{G}'}\neq Y_{\overline{G}''}|\Psi(\overline{G}')=\Psi(\overline{G}'')=g_{exp})\leq 2\zeta.\]

%  Let $err_{\mathcal{O}}(\mathcal{T})=\frac{1}{m}\sum_{(\overline{G},Y)\in \mathcal{T}}\sum_{\overline{G}'\in \mathcal{T}''} \mathds{1}(\widetilde{f}(\overline{G}')\neq Y_{\overline{G'}} \land  \overline{G}'\in S^{\gamma}(\overline{G}))$. From Definition \ref{def:pert_aware} and using linearity of expectation, we have 
% Let $err_{\mathcal{O}}(\mathcal{T})=\frac{1}{m}\sum_{(\overline{G},Y)\in \mathcal{T}''}\mathds{1}(\widetilde{f}(\overline{G})\neq Y_{\overline{G}} \land  \mathcal{F}_{\overline{G}})$, where $\mathcal{F}_{\overline{G}}$ is the event that there exists $\overline{G}'\in \mathcal{T}$ such that $\overline{G}\in S^{\gamma}(\overline{G}')$. From Definition \ref{def:pert_aware} and using linearity of expectation, we have 
% \[\mathbb{E}(err_{\mathcal{O}}(\mathcal{T}))
% =\frac{1}{m}\mathbb{E}(\sum_{(\overline{G},Y)\in \mathcal{T}''}\mathds{1}(\widetilde{f}(\overline{G})\neq Y_{\overline{G}} \land  \mathcal{F}_{\overline{G}}))
% \leq \frac{1}{m}|\mathcal{T}''|P(Y_{\overline{G}}\neq Y_{\overline{G}'}|\mathcal{F}_{\overline{G}})\leq  2\zeta.\]
Consequently, from Hoeffding's inequality, we have $P(err_{\mathcal{O}}\geq 4\zeta)\leq 
2^{-m\zeta^2}$. So, 
\begin{align*}
   &P(\exists f\in \mathcal{H}: \overline{M}_{\mathcal{S}'_1}(\widetilde{f})\geq \overline{M}_{\mathcal{S}_1}(\widetilde{f})+\frac{\epsilon}{4})
   \\&\leq 
   P(\exists f\in \mathcal{H}: \overline{M}_{\mathcal{S}'_1}(\widetilde{f})\geq \overline{M}_{\mathcal{S}_1}(\widetilde{f})+\frac{\epsilon}{4}, err_{\mathcal{O}}\leq 4\zeta)+
   P(err_{\mathcal{O}}\geq 4\zeta)
   \\& \leq 
   P(\exists f\in \mathcal{H}: \overline{M}_{\mathcal{S}'_1}(\widetilde{f})\geq \overline{M}_{\mathcal{S}_1}(\widetilde{f})+\frac{\epsilon}{4}, err_{\mathcal{O}}\leq 4\zeta)+
   2^{-m\zeta^2}
   \\&\leq 
    P(\exists f\in \mathcal{H}: \overline{M}_{\mathcal{S}'_1}(\widetilde{f})\geq \frac{\epsilon}{4}, err_{\mathcal{O}}\leq 4\zeta)+
   2^{-m\zeta^2}
\end{align*}
Let $\mathcal{A}$ be the set of indices of $\mathcal{O}(\overline{G}^{(i)}|\mathcal{T}_i)$ which contain at least one sample which is missclassified by $\widetilde{f}$. Note that since $err_{\mathcal{O}}(\mathcal{T})\leq 4\zeta$, at most $4\zeta m$ of the elements in $\cup_{i\in [\ell]}\mathcal{O}(\overline{G}^{(i)}|\mathcal{T}_i)$ can be in $\mathcal{T}$ and the rest must be in $\mathcal{T}'$. 
Since $\mathcal{T}$ and $\mathcal{T}'$ are generated identically, each element of $\mathcal{T}''$ is in $\mathcal{T}$ or $\mathcal{T}'$ with equal probaility, i.e., with probability equal to $\frac{1}{2}$. 

Let $\mathcal{I}\subseteq [\ell]$ be the set of indices of $\mathcal{O}(\overline{G}^{(i)}|\mathcal{T}_i)$ which have at least one missclassified element. If $|\mathcal{I}|=i$, then  $|\cup_{j\in \mathcal{I}}\mathcal{O}(\overline{G}^{(j)}|\mathcal{T}_j)|\geq \max(i\log^2{m},\frac{m\epsilon}{4})$, by construction. The probability that at most $4\zeta m$ of these elements are in $\mathcal{T}$  is upper bounded by:
\begin{align*}
&P(\exists f\in \mathcal{H}: \overline{M}_{\mathcal{S}'_1}(\widetilde{f})\geq \frac{\epsilon}{4}, err_{\mathcal{O}}\leq 4\zeta)
\\&=
P(\big|\cup_{j\in \mathcal{I}}\mathcal{O}(\overline{G}^{(j)}|\mathcal{T}_j)\cap \mathcal{T}\big|\leq 4\zeta m,\big|\cup_{j\in \mathcal{I}}\mathcal{O}(\overline{G}^{(j)}|\mathcal{T}_j)\cap \mathcal{T}'\big|\geq \frac{m\epsilon}{4} )
    \\&\stackrel{(a)}{\leq} 
    \sum_{i=1}^{\ell} {\frac{2m}{\log^2{m}} \choose i}\sum_{j=1}^{4\zeta m}{\max(i\log^2 m,\frac{m\epsilon}{4}) \choose j} 2^{-\max(i\log^2{m},\frac{m\epsilon}{4})}
    \\&
    \leq \sum_{i=1}^{\ell} {\frac{2m}{\log^2{m}} \choose i}4\zeta m {\max(i\log^2 m,\frac{m\epsilon}{4}) \choose 4\zeta m} 2^{-\max(i\log^2{m},\frac{m\epsilon}{4})}
    \\&\leq 
    \sum_{i=1}^{\ell} {\frac{2m}{\log^2{m}} \choose i}  2^{-\max(i\log^2{m},\frac{m\epsilon}{4})+4\zeta m \log{\max(i\log^2 m,\frac{m\epsilon}{4})} + \log{m}}
    \\& 
    = \sum_{i\in[1,\frac{m\epsilon}{4\log^2{m}}]}  {\frac{2m}{\log^2{m}} \choose i}  2^{-\frac{m\epsilon}{4}+4\zeta m \log{\frac{m\epsilon}{4}} + \log{m}}+
    \sum_{i\in[\frac{m\epsilon}{4\log^2{m}},\ell]}  {\frac{2m}{\log^2{m}} \choose i}  2^{-i\log^2{m}+4\zeta m \log{(i\log^2{m})} + \log{m}}
    \\& \stackrel{(b)}{\leq} \frac{m\epsilon}{4\log^2{m}}{\frac{2m}{\log^2{m}} \choose \frac{m\epsilon}{4\log^2{m}}}  2^{-\frac{m\epsilon}{8}}+\ell \max_{i\in [\ell]}  {\frac{2m}{\log^2{m}} \choose i}  2^{-\frac{1}{2}i\log^2{m}},
\end{align*}
where in (a) we have used the union bound and in (b) we have used the fact that $\epsilon \geq 32 \zeta$. Consequently, 
\begin{align*}
  &  P(\exists f\in \mathcal{H}: \overline{M}_{\mathcal{S}'_1}(\widetilde{f})\geq \overline{M}_{\mathcal{S}_1}(\widetilde{f})+\frac{\epsilon}{4})\leq 
    \frac{m\epsilon}{4\log^2{m}}{\frac{2m}{\log^2{m}} \choose \frac{m\epsilon}{4\log^2{m}}}  2^{-\frac{m\epsilon}{8}}+\ell \max_{i\in [\ell]}  {\frac{2m}{\log^2{m}} \choose i}  2^{-\frac{1}{2}i\log^2{m}}
    \\& \leq 2^{\frac{-m\epsilon}{16}}+ \frac{2m}{\log^2{m}} \max_{i\in [\ell]} 2^{\frac{-1}{2}i \log^2{m}+i\log{2m}}
    \leq 2^{\frac{-m\epsilon}{16}}+\frac{2m}{\log^2{m}} 2^{-\frac{m\epsilon}{32}\log^2{m}} \leq 2^{-\frac{m\epsilon}{32}}.
\end{align*}

So, 
\begin{align*}
&P(\exists f\in \mathcal{H}: \overline{M}_{\mathcal{S}'_1}(\widetilde{f})\geq \frac{\epsilon}{4}, err_{\mathcal{O}}\leq 4\zeta)
   \leq 2^{\frac{-\epsilon m}{32}}+
   2^{-m\zeta^2}\leq 2\cdot 2^{\frac{-\epsilon m}{32}},
\end{align*}
where we have used the fact that $1\geq \epsilon \geq 32 \zeta$.
\\\textbf{Step 2:} Finding an upper-bound for the term $P(\exists f\in \mathcal{H}: \overline{M}_{\mathcal{S}'_2}(\widetilde{f})\geq \overline{M}_{\mathcal{S}_2}(\widetilde{f})+\frac{\epsilon}{4})$:
\\By definition of $VC_{EA}(\mathcal{H},\Psi)$, the number of points in $\mathcal{R}_2$ which can be shattered by $\mathcal{H}$ is at most $d\log^2{m}$, where $d\triangleq VC_{EA}(\mathcal{H},\Psi)$. Let $\mathcal{K}$ be the set of all possible ways to labeling $\mathcal{T}''$ by $\mathcal{H}$. Then,   $|\mathcal{K}|\leq \sum_{i=0}^{d\log^2{(m)}}{2m\choose i}\leq (\frac{2em}{d})^{d\log^{2}(m)}$ by Sauer's lemma. On the other hand:
\begin{align*}
  &  P(\exists f\in \mathcal{H}: \overline{M}_{\mathcal{S}'_2}(\widetilde{f})\geq \overline{M}_{\mathcal{S}_2}(\widetilde{f})+\frac{\epsilon}{4})
  \leq \sum_{K\in \mathcal{K}}P( \overline{M}_{\mathcal{S}'_2}\geq \mathbb{E}(\overline{M}_{\mathcal{S}'_2})+\frac{\epsilon}{8}\text{ or } \overline{M}_{\mathcal{S}_2}\leq \mathbb{E}(\overline{M}_{\mathcal{S}_2})-\frac{\epsilon}{8}|K) 
  \\&\leq (\frac{2em}{d})^{d\log^{2}(m)}(P( \overline{M}_{\mathcal{S}'_2}\geq \mathbb{E}(\overline{M}_{\mathcal{S}'_2})+\frac{\epsilon}{8}|K)+P(\overline{M}_{\mathcal{S}_2}\leq \mathbb{E}(\overline{M}_{\mathcal{S}_2})-\frac{\epsilon}{8}|K) )
  \\&\stackrel{(a)}{\leq} 2(\frac{2em}{d})^{d\log^{2}(m)}e^{-2m(\frac{\epsilon}{8}^2)}\leq e^{-\frac{m\epsilon^2}{32}+d\log^2{(m)}\ln{\frac{2em}{d}}},
\end{align*}
where we have used Hoeffding's inequality in (a). 
Taking $m> \frac{32}{\epsilon^2}\left(d\log^2{(m)}ln(\frac{2em}{d})+ln(\frac{8}{\delta})\right)+\frac{32}{\epsilon}log(\frac{8}{\delta})$, we get $P(\mathcal{E}_{\mathcal{T},\frac{\epsilon}{2}})\leq 2P(\mathcal{E}_{\mathcal{T},\mathcal{T}',\frac{1}{2}\epsilon})\leq 2(\frac{\delta}{8}+\frac{\delta}{8})= \frac{\delta}{2}$. Let ${f}^*$ be the statistically optimal graph classifier in $\mathcal{H}$, and let $\widetilde{f}$ be the EA-ERM output, and let the corresponding classifier minimizing the empirical risk on the augmented dataset be denoted by $f$. Then, 
\begin{align}
  &  P(err_{P_G}(\widetilde{f})\geq err_{P_G}(f^*)+\epsilon)
  \\& \leq 
    P(err_{P_G}(\widetilde{f})\geq err_{\mathcal{T}}(f)+\frac{1}{2}\epsilon \text{ or } err_{\mathcal{T}}(f)> err_{\mathcal{T}}(f^*)\text{ or } err_{\mathcal{T}}(f^*)\geq err_{P_G}(f^*)+\frac{1}{2}\epsilon) 
    \\& {\leq} 
    P(err_{P_G}(\widetilde{f})\geq err_{\mathcal{T}}(f)+\frac{1}{2}\epsilon)
    +
    P(err_{\mathcal{T}}(f)> err_{\mathcal{T}}(f^*))
+P( err_{\mathcal{T}}(f^*)\geq err_{P_G}(f^*)+\frac{1}{2}\epsilon) 
\label{eq:sc0}
\\&\leq P(\mathcal{E}_{\mathcal{T},\frac{\epsilon}{2}})+0+\frac{\delta}{2}\leq \frac{\delta}{2}+ \frac{\delta}{2}=\delta,
\label{eq:sc}
\end{align}
where in \eqref{eq:sc0} we have used the union bound, and in \eqref{eq:sc}
we have used Hoeffding's inequality to conclude that  $P( err_{\mathcal{T}}(f^*)\geq err_{P_G}(f^*)+\frac{1}{2}\epsilon) \leq \frac{\delta}{2}$ and the definition of EA-ERM to conclude that $P(err_{\mathcal{T}}(f)> err_{\mathcal{T}}(f^*))=0$. Consequently,
\begin{align*}
    m_{EA}(\epsilon,\delta,\zeta; \mathcal{H},\Psi) = {O}\left(\frac{d}{\epsilon^2}\log^2{d}+\frac{1}{\epsilon^2}ln(\frac{1}{\delta})\right)
\end{align*}

Note that a lower-bound on sample complexity which is tight with respect to the above upper-bound follows by standard arguments. We provide an outline of the proof in the following. Let us take $\zeta=0$, and let the instance space be such that for each element of the image of the explanation function, there is only one sample with non-zero probability such that the set of probable inputs are shattered by $\mathcal{H}$. Note that such a set of probable inputs always exists by definition of $VC_{EA}(\mathcal{H},\Psi)$. In that case, it is straightforward to see that the sample complexity is the same for generic and explanation-assisted learning rules (since the explanation does not provide any additional information as for each explanation, there is only one probable input) and the lower-bound follows. 
\end{proof}

\section{Detailed Experimental Setup}
\label{sec:app:fullsetup}
Our experiments were conducted on a Linux system equipped with eight NVIDIA A100 GPUs, each possessing 40GB of memory. We use CUDA version 11.3, Python version 3.7.16, and Pytorch version 1.12.1.

\subsection{Datasets}
In our empirical experiments,  we use a benchmark synthetic dataset and 6 real-life datasets. 
\begin{itemize}[leftmargin=*]
    \item \textbf{\bamo}~\cite{luo2020parameterized} dataset includes 1,000 synthetic graphs created from the basic Barabasi-Albert (BA) model. This dataset is divided into two different categories: half of the graphs are associated with `house 'motifs, while the other half are integrated with five-node circular motifs. The labels of these graphs depends on the specific motif they incorporate.
    \item  \textbf{\mutag}~\cite{debnath1991structure} dataset comprises 2,951 molecular graphs, divided into two classes according to their mutagenic effects on the Gram-negative bacterium S. Typhimurium. Functional groups $NO_2$ and $NH_2$ are considered as ground truth explanations for positive samples~\cite{luo2020parameterized}.
    \item \textbf{\benz}~\cite{benjam2020evaluating} is a dataset of 12,000 molecular graphs from the ZINC15 database\cite{teague2015zinc}. The graphs are divided into two classes based on whether they have a benzene ring or not. If a molecule has more than one benzene ring, each ring is a separate explanation.\\
    \item  \textbf{\fluo}~\cite{benjam2020evaluating} dataset contains 8,671 molecular graphs, divided into two classes based on whether they have both a fluoride and a carbonyl group or not. The ground truth explanations are based on the specific combinations of fluoride atoms and carbonyl functional groups found in each molecule.\\
    \item \textbf{\alk}~\cite{benjam2020evaluating} is a dataset of 4,326 molecular graphs, divided into two classes. A positive sample is a molecule with an unbranched alkane and a carbonyl group.     
    \item  \textbf{\dd}~\cite{dobson2003distinguishing}  comprises 1,178 protein structures. proteins are depicted as graphs where each node represents an amino acid. Nodes are interconnected by an edge if the amino acids are within 6 Angstroms of each other. Protein structures into binary classes: enzymes and non-enzymes.
    \item \textbf{\prot}~\cite{dobson2003distinguishing,borgwardt2005protein} consists of 1,113 protein graphs which are generated in the same way as {\dd}. 
\end{itemize}

The statistics of datasets are shown in Table~\ref{tab:datasets}. The \# of explanations denotes the number of graphs with ground truth explanations. 
\begin{table*}[h]
    \centering
        \caption{The detailed information of graph datasets}
    \label{tab:datasets}
    \scalebox{1.00}{
     \begin{tabular}{lrrrrr}
        \hline
        Dataset & \#graphs & \#nodes & \#edges & \#explanations & \#classes \\
        \hline
        \bamo        & 1,000 & 25     & 50-52   & 1,000 & 2   \\
        \mutag       & 2,951  & 5-417  & 8-224   & 1,015 & 2  \\
        \benz        & 12,000 & 4-25   & 6-58    &  6,001 & 2  \\
        \fluo        & 8,671  & 5-25   & 8-58    & 1,527 & 2 \\
        \alk         & 4,326  & 5-25   & 8-58    & 375 & 2 \\
        \dd          & 1,178  &   30-5,748 & 126 -28,534      &   0 & 2 \\
        \prot        & 1,113  &   4-620  &   10-2,098 &   0 & 2 \\
         \hline
\end{tabular}}
\end{table*}

\subsection{GNN Models} 
We use the same GCN model architectures and hyperparameters as~\citep{luo2020parameterized}. Specifically, For the GCN model, we embed the nodes with two GCN-Relu-BatchNorm blocks and one GCN-Relu block to learn node embeddings.  Then, we adopt readout operations~\cite{xu2018powerful} to get graph embeddings, followed by a linear layer for graph classification. The number of neurons is set to 20 for hidden layers. For the GIN model, we replace the GCN layer with a Linear-Relu-Linear-Relu GIN layer.  For the PNA model, we adopt a similar architecture in ~\cite{miao2022interpretable}. We initialize the variables with the Pytorch default setting and train the models with Adam optimizer with a learning rate of $1.0 \times 10^{-3}$. 

\subsection{Data Augmentation Baselines}
\begin{itemize}[leftmargin=*]
    \item Edge Inserting: We randomly select 10\% unconnected node pairs to generate the augmentation graph.
    \item Edge Dropping:  We generate a graph by randomly removing 10\% edges in the original graph.
    \item Node Dropping: We generate a graph by randomly dropping 10\% nodes from the input graph, together with their associated edges 
    \item Feature Dropping: We generate a graph by randomly dropping 10\% features.
    \item Mixup: Given a labeled graph $(\overline{G}_i,Y_i)$, we randomly sample another labelled graph $(\overline{G}_j,Y_j)$. There adjacency matrices are denoted by $\mA_i$ and $\mA_j$, respectively. We generate a block diagonal matrix $\text{diag}(\mA_i,\mA_j)$:  
\begin{equation}
\label{eq:align-adj}
    \text{diag}(\mA_i,\mA_j) = \left[ \begin{array}{cc}
\mA_i & 0 \\ 0 & \mA_j \end{array} \right].
\end{equation}
The corresponding graph is denoted as $\overline{G}^{(\text{diag})}$ We obtain the mixup augmentation graph by randomly adding two cross-graph edges, i.e., one node from $\overline{G}_i$ and the other from $\overline{G}_j$, to $\overline{G}^{(\text{diag})}$.
\end{itemize}

\section{Extra Experiments}
In this section, we provide extensive experiments to further verify the effectiveness of our method and support our theoretical findings.
\label{sec:app:extraexp}
\subsection{Analysis on Distribution of Our Method}
\begin{figure}[h]
  \centering
    \subfigure[ Aug$_\text{GE}$ on \fluo]{\includegraphics[width=0.25\textwidth]{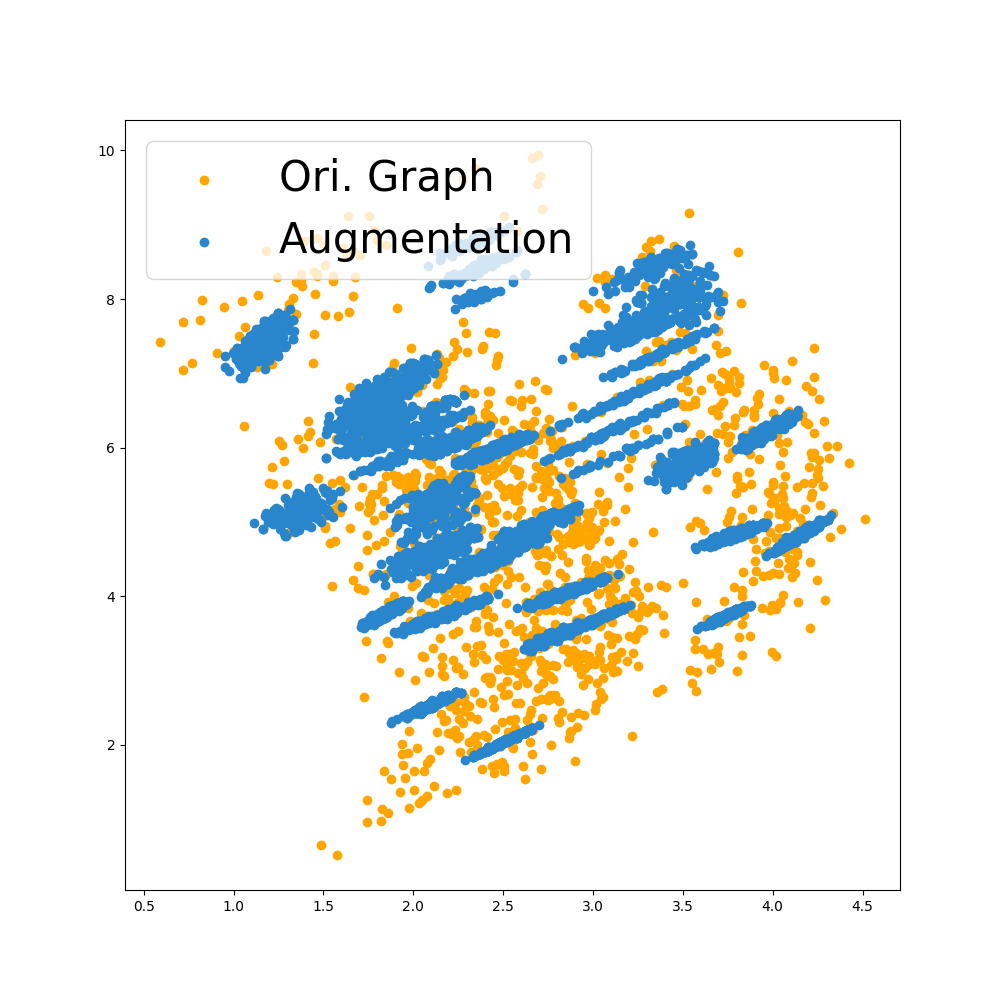} \hspace{-0.9em}\label{fig:exp1:distribution:gnnfluoride}}    
    \subfigure[ Aug$_\text{GE}$ on \alk]{\includegraphics[width=0.25\textwidth]{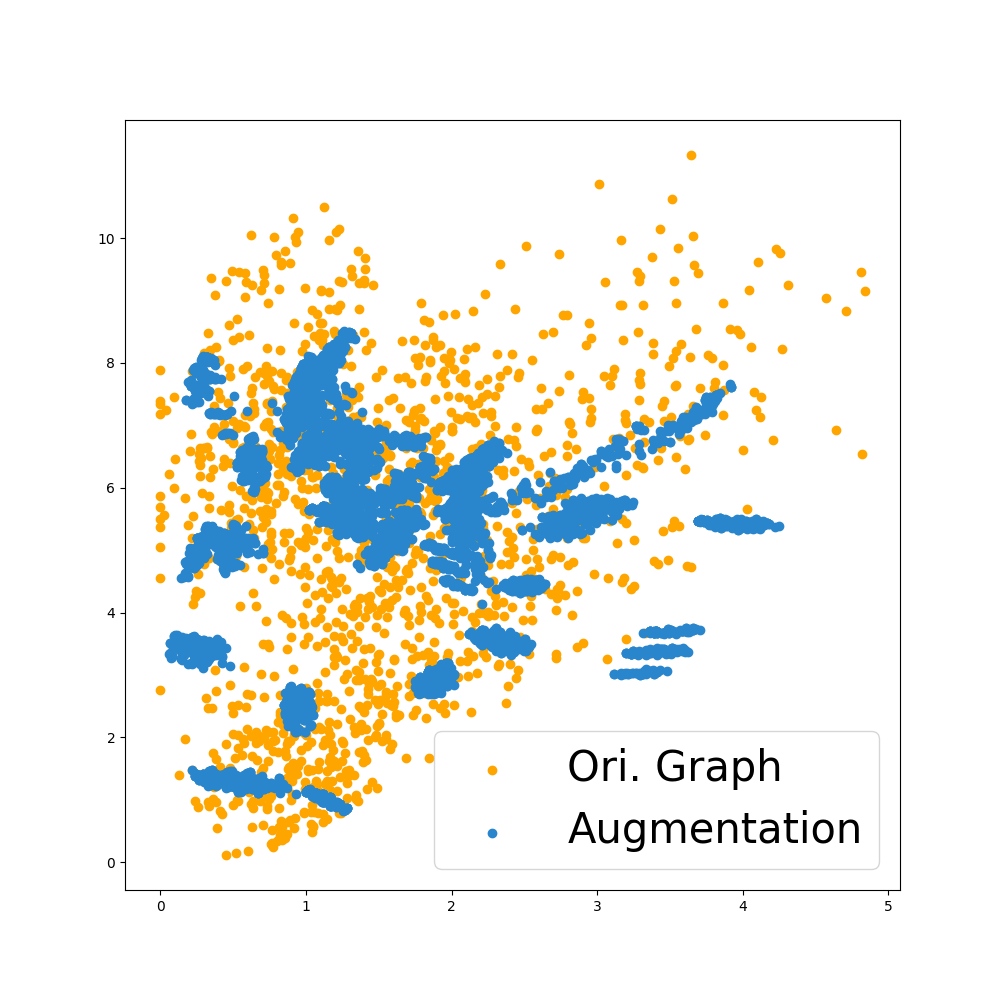}\hspace{-0.9em} \label{fig:exp1:distribution:gnnalkane}} 
    \subfigure[Aug$_\text{PE}$ on \fluo]{\includegraphics[width=0.25\textwidth]{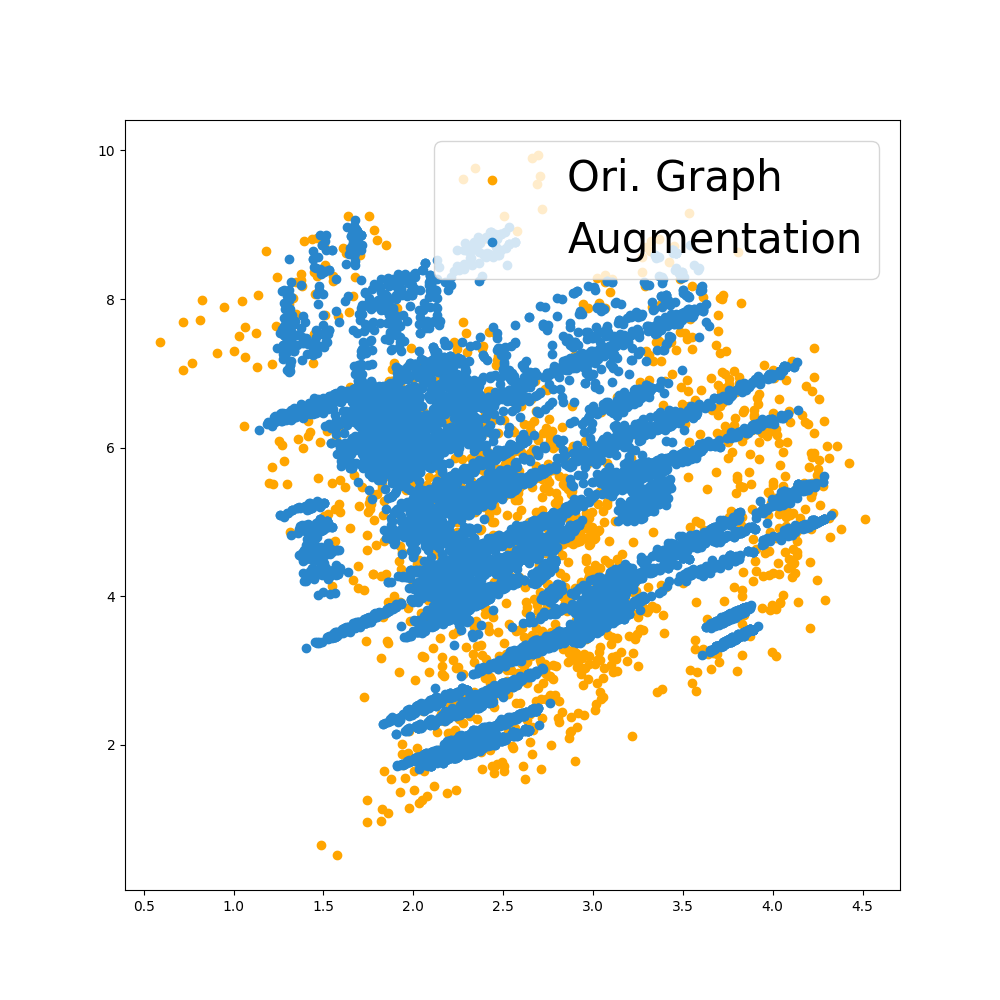} \hspace{-0.9em}\label{fig:exp1:distribution:pgefluoride}}    
    \subfigure[Aug$_\text{PE}$ on \alk]{\includegraphics[width=0.25\textwidth]{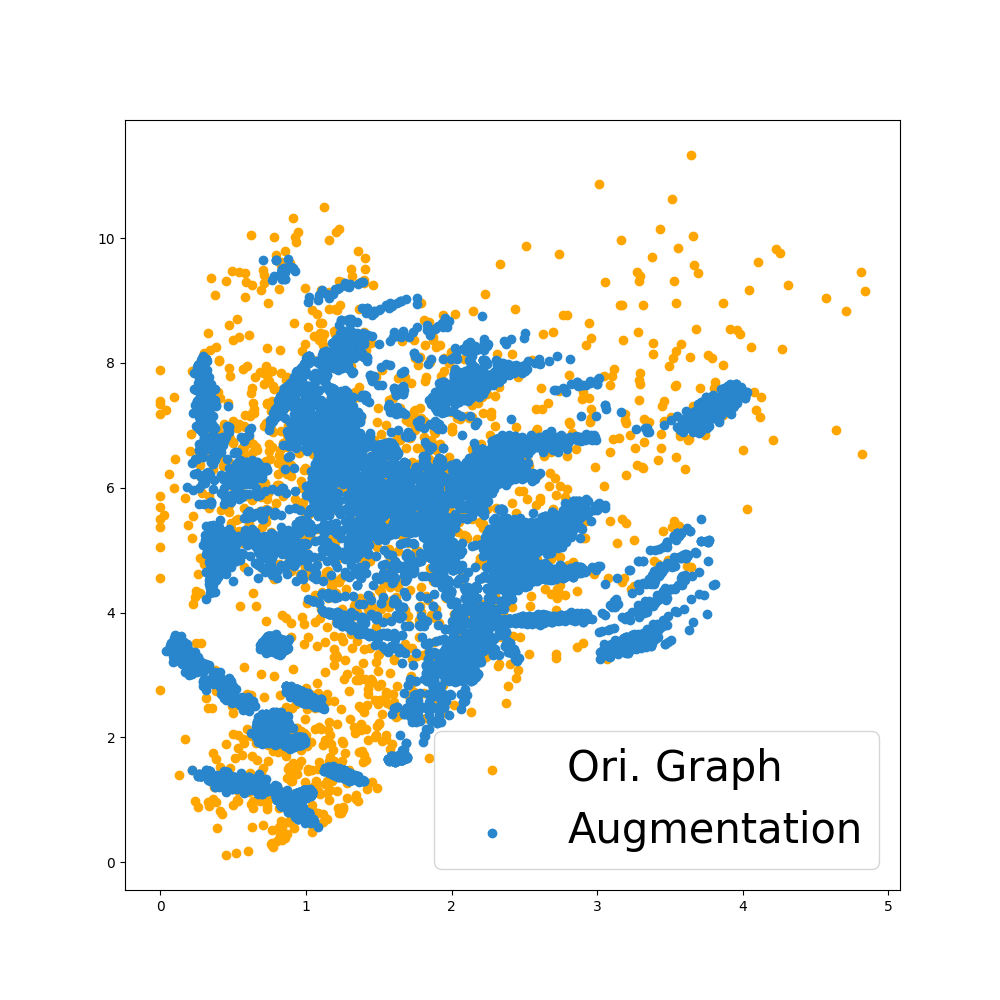}\label{fig:exp1:distribution:pgealkane}} 
    \caption{Visualization results of augmentations generated by Aug$_\text{PE}$ and Aug$_\text{PE}$ (best viewed in color).}
    \label{fig:app:distribution2}
\end{figure}

As described in Section~\ref{sec:learn}, we generate augmentations by an in-distributed perturbation function $\Pi(\cdot)$ (Algorithm~\ref{alg:perturb}). In this part, we empirically verify the effectiveness of our implementation in generating in-distributed augmentations. We use both GNNExplainer~\cite{ying2019gnnexplainer} and PGExplainer~\cite{luo2020parameterized} to generate explanations. Two real-life datasets, {\fluo} and {\alk} are utilized here.
For each dataset, we first pad each graph by inserting isolated nodes such that all graphs have the same size of nodes. Then, for each graph, we concatenate its adjacency matrix with the node matrix followed by a flatten operation to get a high-dimensional vector. We adopt an encoder network to embed high-dimensional vectors into a 2-D vector space. The encoder network consists of two fully connected layers, the same as the decoder network. Cross Entropy is used as the reconstruction error to train the Autoencoder model. The original graphs and augmentations are used for training. The visualization results of these original and augmentation graphs are shown in Figure~\ref{fig:app:distribution2}. We observe that augmentation graphs are in-distributed in both datasets.

\subsection{Dealing with OOD Augmentations}
\begin{figure}[h]
  \centering
    \includegraphics[width=0.40\textwidth]{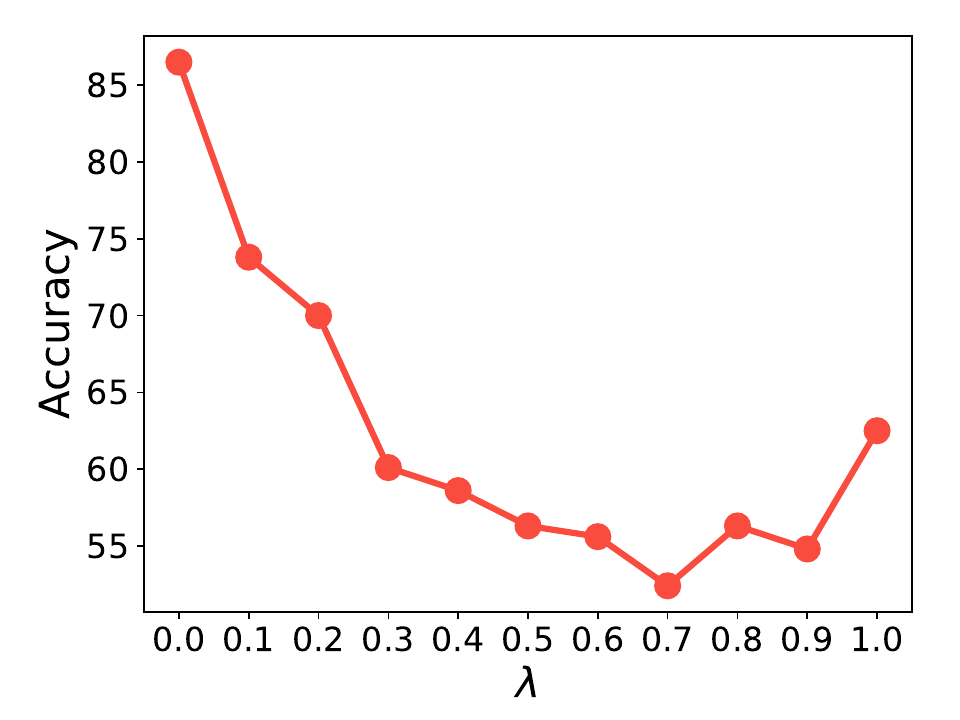} 
    \label{fig:exp1:distribution:ood}
    \caption{ The effects of $\lambda$ in tackling OOD augmentations on \bamo~ dataset.}
\end{figure}

In this section, we conduct experiments to verify the effectiveness of our strategy that includes a hyperparameter $\lambda$ in alleviating the negative effects of OOD graph augmentations. We select $500$, $100$, and $100$ samples in \bamo~ dataset as the training set, valid set, and test set, respectively. To obtain OOD augmentations, we add edges to the non-explanation subgraphs until the average node degree is not less than $17$. Each training instance has $2$ augmentations, and we use $3$ layers GCN as the backbone. As Figure \ref{fig:exp1:distribution:ood} shows, in general, the accuracy decreases as the hyperparameter $\lambda$ rises. The results show that with out-of-distribution graph augmentations, a small $\lambda$ can alleviate the negative effects.

\subsection{Hyper-parameter Sensitivity Studies}
In this section, we show the robustness of our method with a set of hyper-parameter sensitivity studies. Two real datasets, {\mutag} and {\benz}, are used in this part. We choose PGExplainer to generate explanations. 

\begin{figure}[h]
  \centering
    \subfigure[\mutag]{\includegraphics[width=0.3\textwidth]{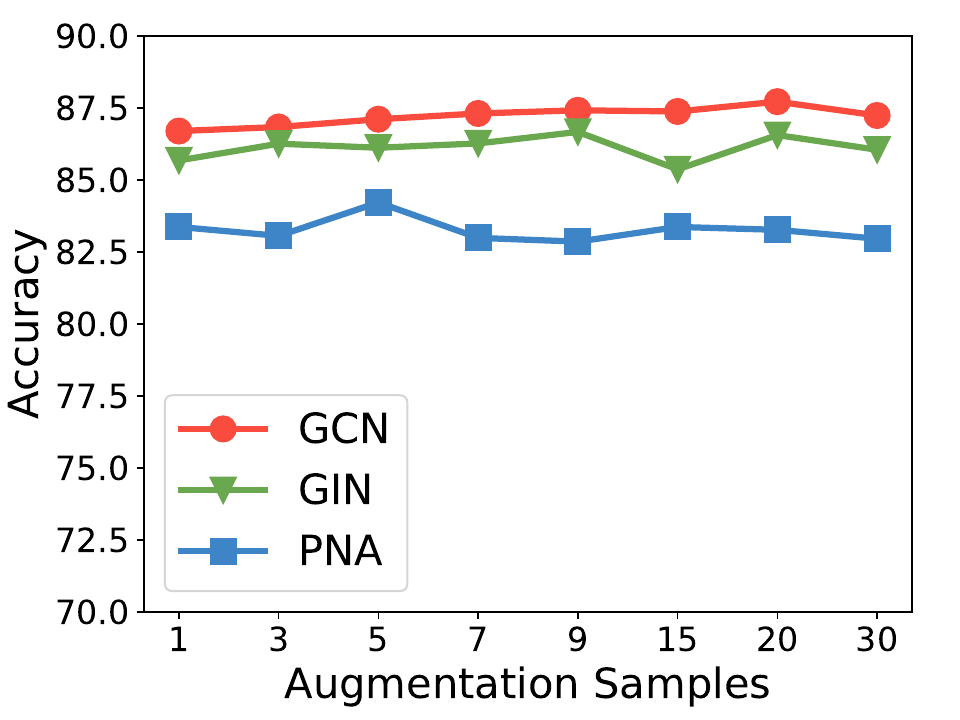}\label{fig:ablation:M:mutag}}    
    \subfigure[\benz]{\includegraphics[width=0.3\textwidth]{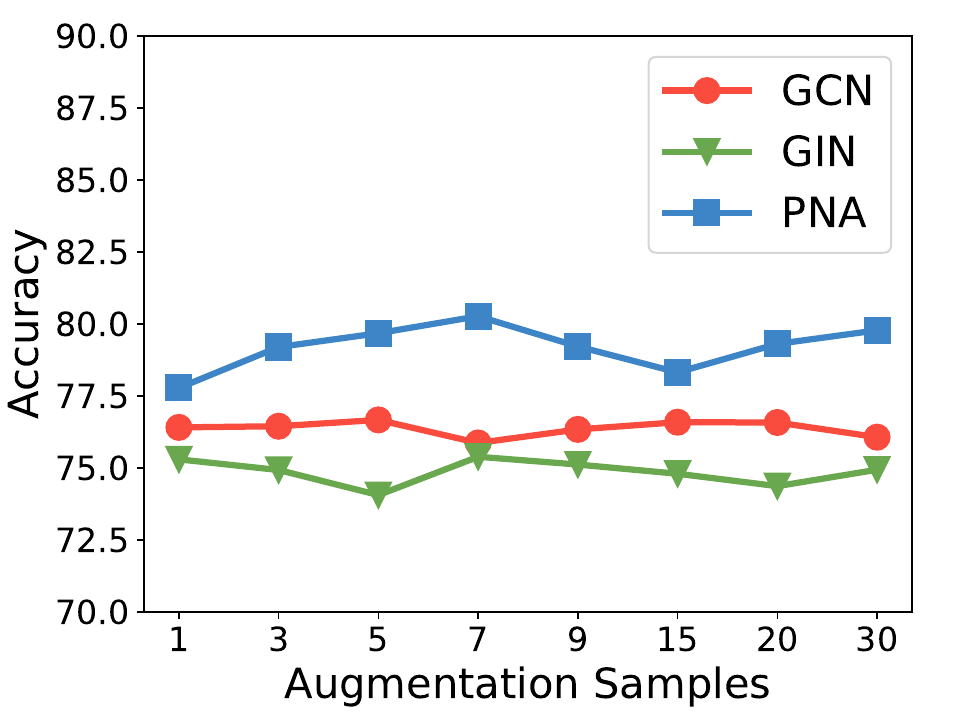}\label{fig:ablation:M:benz}} 
    \caption{Hyper-parametey study of sampling number $M$ by using PGExplainer.}
    \label{fig:app:exp:M}
\end{figure}

As shown in Algorithm~\ref{alg:training}, $M$ denotes the number of augmentation samples per instance. We range the values of $M$ from 1 to 30 and show the accuracy performances of GNN models in Figure~\ref{fig:app:exp:M}. Our method is robust to the selection of $M$. 

% In Eq.~\ref{eq:align-adj}, $\lambda$ is included to avoid the domination of (potentially out-of-distribution) augmented data. 

\subsection{Comparison to Baseline Data Augmentations with 3 layer PNA.}
In this part, we provide the comparison of our methods to baseline data augmentations with the 3-layer PNA as the classifier. As shown in Table~\ref{tab:accuracy:pna3}. we have similar observations with results on GCN and GIN. Our method consistently outperforms other baselines. Specifically, the improvements of Aug$_\text{GE}$ and Aug$_\text{PE}$  over the best of others are 1.99\% and 1.01\%, respectively. 

\begin{table*}[!tp]
\centering
\caption{Performance comparisons with  3-layer PNA trained on 50 samples. The metric is classification accuracy. The best results are shown in bold font and the second best ones are underlined.}
\label{tab:accuracy:pna3}
\begin{tabular}{ll|ccccccccccc}
    \toprule
    \multicolumn{2}{c|}{Dataset} &
    \multicolumn{1}{c}{\mutag} &
    \multicolumn{1}{r}{\benz} & 
    \multicolumn{1}{r}{\fluo} & 
    \multicolumn{1}{r}{\alk} &
    \multicolumn{1}{r}{\dd}&
    \multicolumn{1}{r}{\prot} \\
 \midrule[0.8pt]
\parbox[t]{2mm}{\multirow{8}{*}{\rotatebox[origin=c]{90}{PNA}}}   
& Vanilla                & \ms{83.61}{4.13} & \ms{76.72}{3.28} & \ms{64.90}{6.88} & \underline{\ms{93.60}{1.72}} & \ms{61.29}{5.38} & \underline{\ms{64.91}{5.62}} \\
& Edge Inserting     & \ms{82.35}{3.31} & \ms{78.70}{2.29} & \ms{64.17}{3.76} & \ms{89.66}{4.55} & \ms{61.81}{4.35} & \ms{61.91}{4.44} \\
& Edge Dropping      & \ms{82.18}{2.68} & \ms{77.72}{1.14} & \ms{62.78}{1.76} & \ms{92.31}{2.94} & \ms{61.64}{6.22} & \ms{60.45}{4.35} \\
& Node Dropping      & \ms{81.60}{2.39} & \ms{78.02}{1.46} & \ms{61.59}{3.15} & \ms{90.77}{3.54} & \ms{62.67}{4.28} & \ms{61.64}{5.30} \\
& Feature Dropping   & \ms{81.46}{3.44} & \ms{79.01}{1.74} & \ms{66.63}{3.62} & \ms{90.94}{2.94} & \ms{62.59}{4.53} & \ms{61.18}{3.35} \\
& Mixup   &\ms{72.21}{4.20}   &\ms{75.05}{0.93} &\ms{61.10}{1.18}  &\ms{89.31}{2.37}   &\ms{57.45}{2.07}   &\ms{53.71}{3.36}\\
& Aug$_\text{GE}$  & \textbf{\ms{84.73}{2.04}} & \textbf{\ms{80.61}{0.65}} & \textbf{\ms{68.72}{1.55}} & \textbf{\ms{94.97}{1.46}} & \textbf{\ms{65.43}{6.44}} & \ms{64.64}{5.21} \\
& Aug$_\text{PE}$   & \underline{\ms{84.15}{2.25}} & \underline{\ms{80.47}{0.82}} & \underline{\ms{68.60}{1.96}} & \ms{93.14}{2.52} & \underline{\ms{63.19}{5.23}} & \textbf{\ms{65.09}{4.62}} \\
\bottomrule
\end{tabular}
\end{table*}

% \begin{table*}[h]
%     \centering
%         \caption{Accuracy performance comparison using 3 layers PNA among datasets with 50 samples. We highligh the best and second performance by bold and underline.}
%     \label{tab:datasets:pna3}
%      \begin{tabular}{lrrrrrr}
%         \hline
%         Methods & \mutag & \benz  & \fluo & \alk  & \dd & \prot \\
%         \hline
%         PNA                 & 83.61±4.13 & 76.72±3.28 & 64.90±6.88 & \underline{93.60±1.72} & 61.29±5.38 & \underline{64.91±5.62} \\
%         Edge Inserting      & 82.35±3.31 & 78.70±2.29 & 64.17±3.76 & 89.66±4.55 & 61.81±4.35 & 61.91±4.44 \\
%         Edge Dropping      & 82.18±2.68 &  77.72±1.14 & 62.78±1.76 & 92.31±2.94 & 61.64±6.22 & 60.45±4.35  \\   
%         % Edge Perturbation &   & 5-417  & 8-224   & 1015 & 2  \\
%         Node Insertion     & 81.12±2.95 & 78.31±1.71 & 66.58±3.56 & 90.40±4.14 & 61.55±4.14 & 60.91±5.36  \\
%         Node Dropping      & 81.60±2.39 & 78.02±1.46 & 61.59±3.15 & 90.77±3.54 & 62.67±4.28 & 61.64±5.30  \\
%         Feature Dropping   & 81.46±3.44 & 79.01±1.74 & 66.63±3.62 & 90.94±2.94 & 62.59±4.53 & 61.18±3.35 \\
%         \hline
%         Aug$_\text{GNNE}$  & \textbf{84.73±2.04} & \textbf{80.61±0.65} & \textbf{68.72±1.55} & \textbf{94.97±1.46} & \textbf{65.43±6.44} & 64.64±5.21 \\
%         Aug$_\text{PGE}$   & \underline{84.15±2.25} & \underline{80.47±0.82} & \underline{68.60±1.96} & 93.14±2.52 & \underline{63.19±5.23}  & \textbf{65.09±4.62} \\
%          \hline
% \end{tabular}
% \end{table*}

\subsection{Comparison to Baseline Data Augmentations with Smaller Training Sizes}
\label{sec:app:smalltrainingsize}
We show the performances of GNNs with smaller training sizes to further verify the effectiveness of our methods in improving data efficiency. We consider training sizes with 10 samples and 30 samples in this part. We use GIN in this part and keep other settings the same as Section~\ref{sec:exp:compare2baselines}.  We also include the default setting with 50 samples for comparison. 

From Table~\ref{tab:accuracy:gin3:smallersize}, we observe that Aug$_\text{GE}$ and Aug$_\text{PE}$ improves the accuracy performances by similar margins with smaller training sizes. Specifically, the improvements are 4.38\% and 4.45\% with 10 training samples, and 5.75\% and 5.83\% training samples.  The results are consistent with Section~\ref{sec:exp:augmentationdistribution}, which further verify the effectiveness of our method in boosting the data efficiency for GNN training.

\begin{table*}[!tp]
\centering
\caption{Performance comparisons with  3-layer GIN trained on 10/30/50(default) samples. The metric is classification accuracy. The best results are shown in bold font and the second best ones are underlined.}
\label{tab:accuracy:gin3:smallersize}
\begin{tabular}{ll|ccccccccccc}
    \toprule
    \multicolumn{2}{c|}{Dataset} &
    \multicolumn{1}{c}{\mutag} &
    \multicolumn{1}{r}{\benz} & 
    \multicolumn{1}{r}{\fluo} & 
    \multicolumn{1}{r}{\alk} &
    \multicolumn{1}{r}{\dd}&
    \multicolumn{1}{r}{\prot} \\
 \midrule[0.8pt]
\parbox[t]{2mm}{\multirow{8}{*}{\rotatebox[origin=c]{90}{10 training samples}}}   
& Vanilla                & \ms{80.14}{5.03} & \ms{60.42}{6.10} & \ms{61.62}{2.65} & {\ms{74.78}{9.24}} & \ms{60.18}{5.89} & {\ms{60.52}{9.50}} \\
& Edge Inserting     & \ms{79.05}{3.85} & \ms{64.37}{4.44} & \ms{60.07}{4.84} & \ms{67.89}{12.85} & \ms{55.45}{6.62} & \ms{58.62}{4.06} \\
& Edge Dropping      & \ms{74.80}{3.97} & \underline{\ms{65.13}{2.64}} & \ms{60.20}{4.10} & \ms{74.65}{10.97} & \ms{59.27}{4.65} & \ms{59.05}{5.96} \\
& Node Dropping      & \ms{76.90}{4.74} & \ms{64.88}{2.63} & \ms{59.87}{4.45} & \ms{80.14}{10.98} & \ms{58.27}{5.07} & \ms{58.71}{7.95} \\
& Feature Dropping   & \ms{75.51}{3.98} & \ms{60.95}{5.27} & \ms{61.27}{5.50} & \ms{67.08}{12.35} & \ms{56.64}{5.16} & \ms{58.97}{6.30} \\
& Mixup   &\ms{74.35}{2.09}   &\ms{51.29}{1.36} &\ms{52.25}{2.12}  &\ms{62.76}{2.51}   &\ms{62.27}{3.60}   &\textbf{\ms{64.66}{3.74}}\\
& Aug$_\text{GE}$  & \underline{\ms{83.47}{2.89}} & \textbf{\ms{69.55}{0.91}} & \textbf{\ms{66.54}{1.96}} & \underline{\ms{83.24}{5.82}} & \textbf{\ms{65.45}{3.47}} & \ms{63.62}{5.48} \\
& Aug$_\text{PE}$   & \textbf{\ms{87.48}{2.26}} & \textbf{\ms{69.55}{0.91}} & \underline{\ms{63.00}{3.49}} & \textbf{\ms{84.54}{6.14}} & \underline{\ms{64.91}{2.34}} & \underline{\ms{63.88}{3.07}} \\
\midrule
\parbox[t]{2mm}{\multirow{8}{*}{\rotatebox[origin=c]{90}{30 training samples}}}   
& Vanilla                & \ms{81.09}{3.58} & \ms{64.37}{6.14} & \ms{66.03}{2.61} & {\ms{81.41}{12.50}} & \ms{62.36}{4.85} & {\ms{65.00}{7.29}} \\
& Edge Inserting     & \ms{82.07}{3.87} & \ms{68.92}{3.07} & \ms{64.92}{3.60} & \ms{86.08}{8.59} & \ms{61.91}{5.63} & \ms{64.40}{4.03} \\
& Edge Dropping      & \ms{80.41}{4.02} & \ms{67.48}{4.05} & \ms{61.12}{4.10} & \ms{88.00}{7.09} & \ms{64.09}{4.01} & \ms{62.41}{5.82} \\
& Node Dropping      & \ms{81.02}{4.90} & \ms{67.93}{3.50} & \ms{60.57}{4.66} & \ms{87.61}{7.39} & \ms{64.36}{3.14} & \ms{64.22}{3.32} \\
& Feature Dropping   & \ms{81.60}{4.25} & \ms{63.88}{5.67} & \ms{64.85}{6.23} & \ms{85.28}{6.12} & \ms{62.09}{5.14} & \ms{63.79}{3.99} \\
& Mixup   &\ms{72.24}{2.55}   &\ms{54.28}{2.06} &\ms{52.06}{3.14}  &\ms{68.31}{3.98}   &\ms{56.00}{2.04}   &\ms{60.95}{3.32}\\
& Aug$_\text{GE}$  & \textbf{\ms{84.90}{1.29}} & \underline{\ms{70.69}{1.66}} & \textbf{\ms{71.56}{4.59}} & \textbf{\ms{94.50}{1.32}} & \textbf{\ms{68.55}{5.64}} & \underline{\ms{69.05}{4.23}} \\
& Aug$_\text{PE}$   & \underline{\ms{84.73}{1.45}} & \textbf{\ms{70.81}{2.30}} & \underline{\ms{71.48}{3.64}} & \underline{\ms{94.28}{1.42}} & \underline{\ms{68.00}{6.26}} & \textbf{\ms{70.17}{3.50}} \\
 \midrule
\parbox[t]{2mm}{\multirow{8}{*}{\rotatebox[origin=c]{90}{50 training samples}}}       
&  Vanilla            & \ms{82.52}{3.71}  & \ms{67.48}{5.93} & \ms{68.55}{5.18} & \ms{85.06}{10.27} & \ms{65.14}{4.26} & \ms{66.45}{4.01} \\
& Edge Inserting      & \ms{82.79}{3.21}  & \ms{71.58}{2.77}  & \ms{66.78}{4.04}  & \ms{87.54}{10.32}  & \ms{64.74}{5.38} & \ms{65.45}{5.82}   \\
& Edge Dropping       & \ms{81.63}{3.65}  & \ms{70.46}{4.34}    & \ms{62.91}{5.06}   & \ms{90.29}{6.39} & \ms{66.72}{3.76} & \ms{62.73}{5.29} \\  
& Node Dropping       & \ms{82.18}{3.99} & \ms{71.31}{2.71}   & \ms{64.86}{4.55}    &  \ms{88.89}{7.00} & \ms{66.72}{2.89} & \ms{65.64}{5.38}  \\
& Feature Dropping    & \ms{82.72}{2.92} & \ms{70.66}{2.80}  & \ms{67.58}{5.12} & \ms{83.09}{11.73} & \ms{68.19}{4.34} & \ms{65.55}{5.00} \\
& Mixup   &\ms{74.52}{1.61}   &\ms{59.00}{3.43} &\ms{51.58}{2.59}  &\ms{65.80}{4.13}   &\ms{58.55}{3.48}   &\ms{62.16}{2.92}\\
&Aug$_\text{GE}$  & \underline{\ms{85.99}{2.41}} & \textbf{\ms{75.41}{0.82}} & \underline{\ms{76.29}{2.05}} & \textbf{\ms{94.89}{1.11}} & \textbf{\ms{69.31}{5.19}} & \textbf{\ms{68.45}{5.86}} \\
&Aug$_\text{PE}$   & \textbf{\ms{86.87}{1.79}} & \underline{\ms{75.39}{1.03}} & \textbf{\ms{76.49}{1.68}} &\underline{\ms{94.77}{1.14}} & \underline{\ms{67.41}{2.75}} & \underline{\ms{68.09}{5.52}} \\
\bottomrule
\end{tabular}
\end{table*}

\subsection{Comparison to Baseline Data Augmentations with 1 Layer GNNs}
\label{sec:app:1layergnn}

In this set of experiments, we analyze the effectiveness of our methods on less powerful GNNs. We reduce the GNN layers to 1 for GCN, GIN, and PNA. Other settings are kept the same as in Section~\ref{sec:exp:compare2baselines}. As the results are shown in Table~\ref{tab:1layerperformance}, our methods with GNNExplainer and PGExplainer occupy the best and second-best positions than other six baselines, respectively. Specifically, our methods achieve $3.69\%, 3.99\%, 4.04\%$ improvements with GNNExplainer and $3.89\%, 3.65\%, 3.27\%$ improvements with PGExplainer on average with GCN, GIN, and PNA backbones. Similar to the results of Section~\ref{sec:exp:compare2baselines}, these results show that our methods can enhance the GNN performance on both powerful and less powerful GNNs.

\begin{table*}[!tp]
\centering
% \fontsize{7}{8}\selectfont  
% \setlength\tabcolsep{2pt}
\caption{Accuracy performance comparison using 1 layer GNNs among datasets with 50 samples. We highlight the best and second performance by bold and underlining.} \label{tab:1layerperformance}
\begin{tabular}{ll|ccccccccccc}
    \toprule
    \multicolumn{2}{c|}{Dataset} &
    \multicolumn{1}{c}{\mutag} &
    \multicolumn{1}{r}{\benz} & 
    \multicolumn{1}{r}{\fluo} & 
    \multicolumn{1}{r}{\alk} &
    \multicolumn{1}{r}{\dd}&
    \multicolumn{1}{r}{\prot} \\
 \midrule[0.8pt]
 \parbox[t]{2mm}{\multirow{8}{*}{\rotatebox[origin=c]{90}{GCN}}}
& Vanilla                 & \ms{79.83}{3.21} & \ms{61.46}{4.39} & \ms{56.77}{4.46} & \ms{94.89}{1.65} & \ms{61.38}{6.70} & \ms{66.45}{6.95} \\
& Edge Inserting      & \ms{79.42}{3.95} & \ms{65.36}{4.91} & \ms{54.84}{3.55} & \ms{92.71}{3.79} & \ms{65.26}{7.58} & \ms{62.18}{4.69} \\
& Edge Dropping       & \ms{78.10}{4.50} & \ms{66.02}{4.13} & \ms{54.92}{3.47} & \ms{94.40}{1.76} & \ms{66.12}{7.20} & \ms{62.18}{4.57} \\
& Node Dropping       & \ms{78.54}{4.24} & \ms{66.19}{4.04} & \ms{54.49}{3.43} & \ms{93.54}{2.96} & \ms{66.21}{6.11} & \ms{63.00}{5.65} \\
& Feature Dropping    & \ms{79.56}{3.80} & \ms{63.92}{4.68} & \ms{54.46}{3.55} & \ms{93.11}{2.79} & \ms{66.55}{3.93} & \ms{62.91}{4.30} \\
& Mixup   &\ms{56.67}{2.11}   &\ms{50.11}{0.42} &\ms{51.94}{0.53}  &\ms{61.71}{0.00}   &\ms{59.09}{0.00}   &\ms{56.12}{0.46}\\
& Aug$_\text{GE}$   & \underline{\ms{84.63}{1.65}} & \underline{\ms{69.63}{2.03}} & \textbf{\ms{61.81}{1.54}} & \underline{\ms{95.69}{1.19}} & \underline{\ms{66.98}{5.38}} & \underline{\ms{66.82}{4.98}} \\
& Aug$_\text{PE}$    & \textbf{\ms{85.17}{1.63}} & \textbf{\ms{69.64}{2.07}} & \underline{\ms{61.34}{1.41}} & \textbf{\ms{95.86}{1.14}} & \textbf{\ms{67.67}{5.82}} & \textbf{\ms{66.91}{4.83}} \\
\midrule
\parbox[t]{2mm}{\multirow{8}{*}{\rotatebox[origin=c]{90}{GIN}}}       
& Vanilla                & \ms{81.39}{1.71} & \ms{63.71}{3.33} & \ms{60.31}{4.52} & \ms{87.60}{5.40} & \ms{66.21}{8.48} & \ms{66.45}{3.85} \\
& Edge Inserting     & \ms{81.39}{2.54} & \ms{65.95}{4.06} & \ms{60.06}{3.21} & \ms{85.46}{9.23} & \ms{65.34}{5.57} & \ms{66.18}{5.22} \\
& Edge Dropping      & \ms{81.29}{1.57} & \ms{66.03}{4.06} & \ms{59.86}{3.66} & \underline{\ms{88.31}{5.95}} & \ms{66.98}{3.97} & \ms{66.09}{4.81} \\
& Node Dropping      & \ms{81.33}{1.84} & \ms{65.58}{3.46} & \ms{60.55}{2.78} & \ms{88.06}{7.20} & \ms{65.78}{4.50} & \ms{67.00}{5.69} \\
& Feature Dropping   & \ms{81.12}{1.78} & \ms{65.87}{3.33} & \ms{60.45}{2.36} & \ms{85.51}{9.41} & \ms{67.07}{4.86} & \ms{62.64}{4.21} \\
& Mixup   &\ms{70.03}{3.99}   &\ms{50.19}{0.31} &\ms{51.26}{1.16}  &\ms{70.29}{0.00}   &\ms{54.18}{3.26}   &\ms{68.53}{0.88}\\
& Aug$_\text{GE}$  & \underline{\ms{82.45}{1.18}} & \underline{\ms{66.52}{1.55}} & \underline{\ms{64.80}{1.28}} & \textbf{\ms{95.09}{0.98}} & \textbf{\ms{72.33}{3.71}} & \underline{\ms{68.09}{4.09}} \\
& Aug$_\text{PE}$   & \textbf{\ms{83.16}{2.10}} & \textbf{\ms{66.55}{1.31}} & \textbf{\ms{65.10}{1.39}} & \textbf{\ms{95.09}{0.98}} & \underline{\ms{69.22}{4.28}} & \textbf{\ms{68.91}{4.18}} \\
\midrule
\parbox[t]{2mm}{\multirow{8}{*}{\rotatebox[origin=c]{90}{PNA}}}   
& Vanilla                & \ms{83.67}{4.78} & \ms{73.74}{4.57} & \ms{60.76}{4.57} & \ms{87.77}{9.73} & \ms{62.07}{3.60} & \ms{67.18}{3.76} \\
& Edge Inserting     & \ms{82.07}{2.68} & \ms{75.66}{2.02} & \ms{59.55}{3.35} & \ms{89.00}{4.40} & \underline{\ms{64.22}{4.58}} & \ms{65.00}{5.71} \\
& Edge Dropping      & \ms{82.48}{2.97} & \ms{74.75}{2.46} & \ms{59.15}{3.71} & \ms{91.74}{3.02} & \underline{\ms{64.22}{5.76}} & \ms{62.00}{8.01} \\
& Node Dropping      & \ms{82.45}{2.36} & \ms{75.11}{1.62} & \ms{58.72}{3.38} & \ms{91.37}{2.54} & \ms{61.81}{7.44} & \ms{63.91}{8.29} \\
& Feature Dropping   & \ms{82.65}{3.27} & \ms{75.60}{2.22} & \ms{61.20}{4.55} & \ms{91.03}{2.43} & \ms{65.26}{3.56} & \ms{64.64}{6.28} \\
& Mixup   &\ms{50.00}{0.00}   &\ms{70.57}{1.30} &\ms{55.30}{4.20}  &\ms{38.29}{0.00}   &\ms{55.45}{2.73}   &\ms{50.52}{3.37}\\
& Aug$_\text{GE}$  & \textbf{\ms{84.39}{1.55}} & \underline{\ms{77.91}{1.54}} & \textbf{\ms{67.32}{2.68}} & \textbf{\ms{94.60}{1.10}} & \ms{63.10}{3.65} & \textbf{\ms{69.36}{4.90}} \\
& Aug$_\text{PE}$   & \underline{\ms{84.35}{1.32}} & \textbf{\ms{78.12}{1.37}} & \underline{\ms{66.41}{2.43}} & \underline{\ms{93.74}{1.12}} & \textbf{\ms{66.03}{3.66}} & \underline{\ms{68.55}{6.03}} \\
\bottomrule
\end{tabular}
\end{table*}

\end{document}